\documentclass[preprint,review,longtitle]{elsarticle}

\usepackage{amsmath,amssymb,amsfonts}
\usepackage{graphicx}
\usepackage{subcaption}
\usepackage{booktabs}
\usepackage{multirow}
\usepackage{textcomp}
\usepackage{hyperref}
\usepackage{array}
\usepackage{tabularx}
\usepackage{ragged2e}
\usepackage{threeparttable}
\usepackage{csquotes}
\usepackage{bbm}
\usepackage{algorithm}
\usepackage{algorithmic}
\usepackage{placeins}
\usepackage{enumitem}

\usepackage{etoolbox}
\pretocmd{\longtable}{\small}{}{}

\usepackage{listings}

\newcolumntype{L}[1]{>{\raggedright\arraybackslash}p{#1}}
\newcolumntype{Y}{>{\RaggedRight\arraybackslash}X}

\begin{document}

\begin{frontmatter}

\title{Learning Preference-Based Objectives from Clinical Narratives for Dynamic Sepsis Treatment
}

\author[1]{Daniel J. Tan}
\author[2]{Jayne Hui Zhen Chan}
\author[2]{Kai Wen Hwang}
\author[2]{Arturo Yong Yao Neo}
\author[2]{Kay Choong See}
\author[3]{Mengling Feng\textsuperscript{*}}

\address[1]{Institute of Data Science, National University of Singapore, Singapore}
\address[2]{National University Hospital, Singapore}
\address[3]{Saw Swee Hock School of Public Health, National University of Singapore, Singapore}

\begin{abstract}

Designing reward functions for reinforcement learning (RL) in healthcare remains challenging because clinically meaningful outcomes are sparse, delayed, and difficult to explicitly specify. Although structured clinical data capture physiologic states, they often fail to reflect broader aspects of patient trajectories such as treatment response, recovery dynamics, and intervention burden. Clinical narratives, by contrast, encode longitudinal clinician assessments of disease progression, treatment effectiveness, and recovery, providing a potential source of trajectory-level supervision beyond predefined outcome metrics. We propose Clinical Narrative-informed Preference Rewards (CN-PR), a framework that learns reward functions directly from discharge summaries by treating clinical narratives as scalable supervision for trajectory-level preferences. Using a large language model, we derive trajectory quality scores and construct pairwise preferences between patient trajectories to learn rewards through preference-based optimization. To account for variability in narrative informativeness, we incorporate a task relevance signal that weights supervision according to its relevance to the downstream decision-making task. We evaluate CN-PR in dynamic sepsis treatment using offline RL. The learned reward demonstrated strong monotonic alignment with trajectory quality scores and produced policies associated with improved recovery-related outcomes, including increased organ support–free days and faster shock resolution, while maintaining mortality performance comparable to outcome-based reward baselines. These findings were preserved under external validation. Our results suggest that clinical narratives provide a scalable and expressive source of supervision for reward learning in dynamic treatment regimes.

\end{abstract}

\begin{keyword}
Reinforcement learning \sep inverse reinforcement learning \sep clinical narratives \sep preference learning \sep critical care \sep dynamic treatment regimes
\end{keyword}

\end{frontmatter}

\vspace{-1em} 
\textsuperscript{*}\textit{Corresponding author: ephfm@nus.edu.sg}

\section{Introduction}

Clinical decision-making in intensive care units (ICUs) involves complex, time-dependent treatment strategies that must balance competing objectives under uncertainty. Management of critically ill patients, particularly in conditions such as sepsis, often requires sequential adjustments to interventions including fluid therapy, vasopressors, mechanical ventilation, sedation, and renal support \cite{cecconi2018sepsis}. These decisions are inherently dynamic, motivating the use of data-driven approaches for learning dynamic treatment regimes (DTRs) from electronic health record (EHR) data.

Reinforcement learning (RL) provides a principled framework for modeling such sequential decision-making problems, where policies are optimized to maximize long-term outcomes. Prior work has demonstrated the potential of RL in critical care settings, including applications in sepsis management, ventilation strategies, and hemodynamic control \cite{komorowski2018artificial, raghu2017continuous, roggeveen2021transatlantic}. However, a central challenge remains reward specification, which defines the objective optimized by the learned policy.

Most clinical RL approaches rely on simplified reward functions based on terminal outcomes (e.g., mortality) or a limited set of physiological proxies. While informative, these signals provide only a partial view of patient trajectories and often fail to capture broader aspects of recovery, treatment burden, and clinical progression. Handcrafted intermediate rewards partially address this limitation but introduce heuristic assumptions and remain highly sensitive to reward design choices, often leading to inconsistent policy behavior \cite{luo2024reinforcement, liang2025methods}. As a result, specifying clinically meaningful reward functions in complex healthcare settings remains fundamentally challenging.

An alternative is to infer reward functions from preferences rather than explicitly defining them. Preference-based approaches learn from comparative evaluations of trajectories \cite{brown2019extrapolating}, allowing reward functions to reflect relative outcome quality instead of fixed endpoints. However, existing methods typically rely on manually collected annotations, which are costly and difficult to scale in healthcare settings.

In contrast, routine clinical practice produces a rich and largely untapped source of implicit preference signals: unstructured clinical narratives. Discharge summaries and clinical notes synthesize clinician assessments of treatment response, disease progression, and recovery \cite{johnson2023mimic, spasic2020clinical}, providing high-level evaluations of patient trajectories that are not captured in structured data alone. These narratives therefore provide naturally occurring supervision, where relative trajectory quality is implicitly encoded in how patient courses are described.

Recent advances in large language models (LLMs) enable the extraction of structured signals from clinical text at scale, making it feasible to operationalize such narratives for reward learning. Rather than relying on handcrafted objectives or explicit outcome labels, preference-based objectives learn reward functions from relative comparisons between patient trajectories. Narrative-derived signals therefore provide a scalable proxy for trajectory-level preferences, capturing broader aspects of clinical trajectories beyond predefined endpoints.

Motivated by this perspective, we propose \textit{Clinical Narrative-informed Preference Rewards} (CN-PR), a framework for learning reward functions directly from clinical narratives. A large language model is used to assign trajectory quality scores (TQS) from discharge summaries, which are subsequently converted into pairwise preferences over patient trajectories.

Our specific contributions include:
\begin{enumerate}
    \item We introduce a method for deriving trajectory-level preference signals from unstructured clinical narratives, together with a task-relevance-weighted supervision mechanism to account for variability in narrative informativeness.
    
    \item We formulate a structured preference-based objective that captures both the ordering and relative magnitude of trajectory-level signals without requiring handcrafted rewards or explicit annotations.
    
    \item We develop a comprehensive regression-based evaluation framework that characterizes policy behavior across multiple dimensions of treatment and recovery quality, enabling systematic comparison with standard reward formulations in clinical RL.
\end{enumerate}

We evaluate the framework in critical care using dynamic sepsis treatment, a canonical sequential decision-making problem in healthcare RL \cite{komorowski2018artificial,luo2024reinforcement}.

\section{Background and Related Work}

\subsection{\textbf{Reinforcement Learning for Dynamic Treatment Regimes}}

Reinforcement learning (RL) has been increasingly applied to dynamic treatment regimes (DTRs), where policies optimize sequential treatment decisions based on patient state \cite{komorowski2018artificial}. Applications include sepsis management, mechanical ventilation, and sedation using retrospective EHR data \cite{raghu2017continuous, peine2021development, eghbali2021patient}. In these settings, learning is performed offline without the ability to safely explore new actions, making policy behavior highly dependent on the reward function used during training. Consequently, reward specification plays a central role in determining both the behavior and clinical plausibility of learned policies.

\subsection{\textbf{Reward Learning and Preference-Based Methods}}

Designing reward functions for clinical RL remains challenging due to the multidimensional nature of clinical outcomes. Most existing approaches rely on terminal endpoints or handcrafted physiological proxies, which can introduce subjective assumptions and lead to unstable or poorly aligned policies \cite{luo2024reinforcement, liang2025methods}.

Inverse reinforcement learning (IRL) provides a data-driven alternative to manual reward design by inferring reward functions from observed behavior \cite{adams2022survey}. In healthcare, IRL has been used to model clinician decision-making in settings such as sepsis treatment and mechanical ventilation \cite{yu2019inverse, yu2019deep}. However, classical IRL assumes that observed behavior is approximately optimal, an assumption that is often violated in clinical practice due to variability in expertise, institutional protocols, and patient heterogeneity \cite{wang2020adversarial, berner2008overconfidence}.

Preference-based IRL relaxes this assumption by learning from comparisons between trajectories rather than demonstrations alone, allowing reward learning from imperfect data \cite{brown2019extrapolating}. While promising, these approaches typically require explicit preference annotations, which are costly and difficult to obtain at scale in clinical settings.

More broadly, our work aligns with perspectives that treat reward functions as imperfect observations of underlying objectives rather than definitive specifications \cite{hadfield2017inverse}, motivating the use of diverse and implicit sources of supervision for reward learning.

\subsection{\textbf{Learning from Language and Clinical Narratives}}

Recent work has demonstrated that reward functions can be learned from human judgments expressed through language, as in reinforcement learning from human feedback (RLHF) \cite{stiennon2020learning, ouyang2022training}. These approaches highlight the potential of language as a scalable medium for specifying complex objectives.

In healthcare, clinical narratives such as discharge summaries provide rich descriptions of patient trajectories, treatment responses, and outcomes. Prior work has demonstrated the value of clinical text for predictive modeling \cite{huang2019clinicalbert}, yet most RL-based DTR frameworks rely primarily on structured data and do not leverage narratives for reward construction.

Large language models (LLMs) enable scalable extraction of structured information from unstructured clinical text, creating an opportunity to use narratives as a source of implicit preference supervision. However, this direction remains largely unexplored in clinical reinforcement learning.

\subsection{\textbf{Gap in Existing Work}}

Existing approaches rely primarily on either manually specified rewards or explicitly collected preference annotations, both of which present limitations in scalability, expressiveness, and clinical alignment. Despite containing rich trajectory-level assessments of treatment response and recovery, clinical narratives remain largely unused for reward construction in RL-based dynamic treatment regimes. This gap motivates our proposed framework, which leverages narrative-derived preference signals to enable scalable and clinically grounded reward learning.

\section{Methods}

Our study is reported in accordance with the TRIPOD+AI guidelines to ensure transparent, reproducible, and comprehensive reporting of clinical prediction models \cite{collins2024tripod+}.

\subsection{\textbf{Data Extraction and Preprocessing}}

We extracted ICU stays from the MIMIC-IV database \cite{johnson2023mimic}, which contains data for over 60{,}000 ICU patients admitted to Beth Israel Deaconess Medical Center between 2008 and 2019. Inclusion criteria were: (1) sepsis defined according to Sepsis-3 criteria, (2) availability of a corresponding discharge summary within 14 days of ICU discharge, (3) selection of the final ICU stay within each hospital admission, and (4) ICU length of stay between 12 hours and 21 days. These criteria were designed to ensure temporal alignment between structured trajectories and narrative summaries while excluding poorly documented or clinically implausible episodes, thereby improving the reliability of narrative-derived supervision for downstream reward learning.

A total of 30{,}594 hospital stays met the initial inclusion criteria for narrative extraction. After additional preprocessing for feature availability and trajectory construction, the final cohort consisted of 25{,}370 ICU stays. The extracted data span from 24 hours before sepsis diagnosis to 48 hours after diagnosis, forming a maximum 72-hour window per patient. Full cohort details are provided in Supplementary Table~S1.

Clinical features spanning demographics, vital signs, laboratory measurements, and interventions were extracted as summarized in Table~\ref{tab:included_variables}. Values exceeding clinically plausible ranges were clipped, and continuous variables were standardized. Patient trajectories were discretized into 4-hour intervals using summary statistics within each interval, following established preprocessing procedures from prior work \cite{komorowski2018artificial, raghu2017continuous}. Missing values were handled using sample-and-hold imputation augmented with time-since-last-measurement features.

Intravenous fluid administration included bolus and continuous infusions of crystalloids, colloids, and blood products, normalized for tonicity \cite{waechter2014interaction}. Vasopressor therapy comprised norepinephrine, epinephrine, vasopressin, dopamine, and phenylephrine, with doses converted to norepinephrine equivalents using established dose-equivalence relationships \cite{brown2013survival}. Additional details regarding action discretization are provided in Supplementary Table~S2.

\subsection{\textbf{Clinical Narrative Extraction and Processing}}

We extracted deidentified discharge summaries from the MIMIC-IV Notes dataset, where each hospital admission is associated with a single discharge note. Each ICU stay was paired with its corresponding narrative.

To focus on clinically meaningful reasoning, we retained sections containing high-level clinical interpretation:
\begin{itemize}
    \item History of Present Illness
    \item Hospital Course
    \item Discharge Diagnosis
    \item Discharge Condition
    \item Discharge Disposition
    \item Discharge Instructions
\end{itemize}

Sections dominated by raw data (e.g., lab tables or imaging reports) were excluded, as they lack contextual interpretation and increase input length without improving outcome inference.

To provide additional context, we prepended structured metadata to each narrative:
\begin{itemize}
    \item Dialysis status at ICU admission and discharge
    \item Baseline SOFA score
    \item ICU length of stay
\end{itemize}

These variables were included to contextualize disease severity without directly determining outcome labels.

\subsection{\textbf{LLM-Based Clinical Trajectory Scoring}}

We employed a large language model (GPT-OSS-120B) to assign trajectory quality scores (TQS) from discharge summaries using a structured prompt designed to emulate expert clinical reasoning. The model was used in a zero-shot setting without task-specific fine-tuning.

Each trajectory was assigned a score $y \in \{1,2,3,4,5\}$ reflecting overall clinical trajectory quality, incorporating not only survival outcomes but also treatment response, complication burden, and recovery dynamics. Importantly, patients with the same survival outcome may experience substantially different clinical courses—for example, two survivors may differ markedly in recovery trajectory, complication burden, and long-term prognosis, while deaths may arise from either rapid deterioration or prolonged treatment-resistant illness. Such distinctions are not captured by binary outcome endpoints, which implicitly treat all survivors or non-survivors as homogeneous despite substantial variation in treatment response and recovery trajectory.

The five-category formulation was designed to balance expressiveness and reliability, providing sufficient granularity to distinguish clinically meaningful trajectory patterns while minimizing ambiguity arising from subtle or poorly specified narrative differences. Although the CN-PR framework can accommodate alternative discretizations, we found this level of granularity to provide a practical balance between resolution and robustness for narrative-derived supervision in the sepsis treatment setting. More generally, the appropriate degree of discretization depends on both the richness of the clinical narratives and the complexity of the underlying treatment decision problem.

\begin{itemize}
    \item \textbf{1:} Deceased following rapid or potentially modifiable deterioration, with limited evidence of stabilization or treatment response
    \item \textbf{2:} Deceased following a prolonged or largely irreversible course, with persistent organ failure despite treatment
    \item \textbf{3:} Survival with poor outcome, characterized by severe unresolved complications, limited recovery, or near-terminal trajectory
    \item \textbf{4:} Survival with partial recovery, where key complications are controlled but significant morbidity or functional limitations remain
    \item \textbf{5:} Survival with favorable recovery, marked by effective treatment response, resolution of acute illness, and minimal persistent dysfunction
\end{itemize}

In addition to TQS, the model outputs a scalar \textit{task relevance score} reflecting how informative each discharge summary is for the target decision-making task. Task relevance is higher when narratives explicitly describe sepsis-related processes (e.g., hemodynamic instability, vasopressor use, fluid resuscitation) and their relationship to clinical outcomes, and lower when such information is sparse or dominated by unrelated conditions. This mechanism downweights trajectories that provide limited or ambiguous supervision and helps align preference learning with the downstream sepsis treatment objective.

For interpretability, the model also generates a brief textual justification supporting each assigned score. To assess scoring reliability, a subset of 150 discharge summaries was independently reviewed by three clinicians, who assigned both TQS and task relevance scores using the same annotation framework. Agreement between clinicians and the LLM, as well as inter-clinician agreement, was assessed using quadratic-weighted Cohen’s $\kappa$, Spearman rank correlation ($\rho$), exact agreement, and within-one-category agreement.

To promote reproducibility, deterministic decoding settings were used to reduce stochastic variation in model outputs under fixed conditions. The complete prompting framework is provided in Supplementary Information Section~1.1, together with a representative discharge summary, assigned scores, and model-generated justification.

\begin{figure*}[h]
\centering
\includegraphics[width=0.7\textwidth]{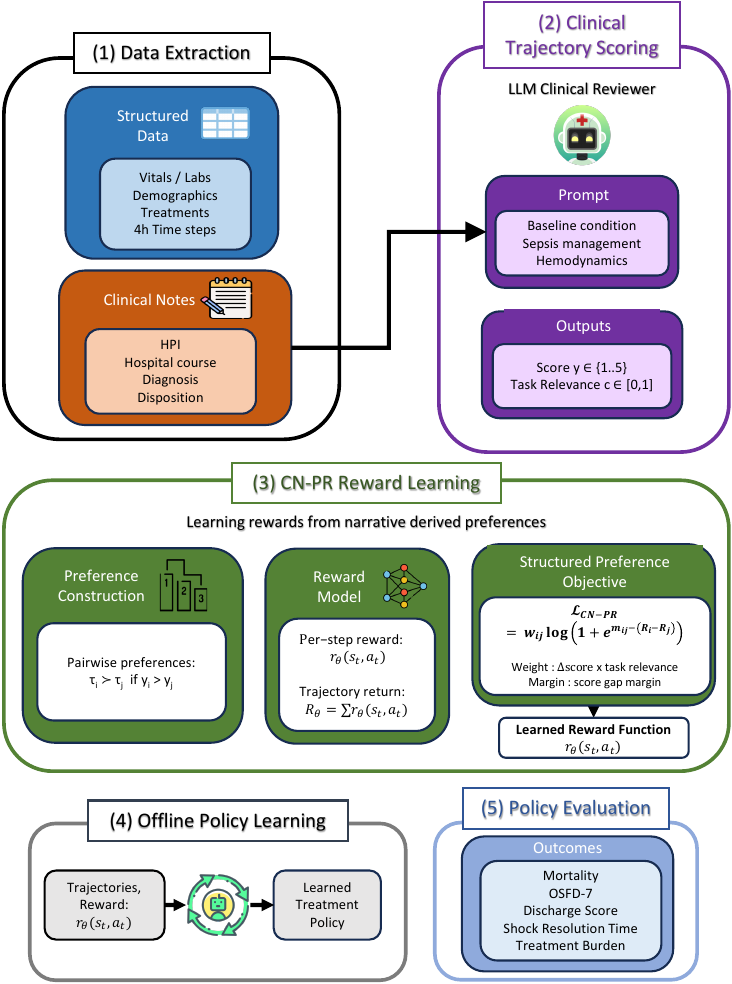}
\caption{
Overview of the CN-PR framework.
}
\label{fig:cnpr_pipeline}
\end{figure*}

An overview of the full CN-PR pipeline is shown in Figure~\ref{fig:cnpr_pipeline}.

\FloatBarrier
\subsection{\textbf{State and Action Space}}

We model clinical decision-making as a Markov Decision Process (MDP) $(\mathcal{S}, \mathcal{A}, P, r, \gamma)$. The state space is continuous and $\mathcal{S}$ comprises 48 clinical variables (Table~\ref{tab:included_variables}), consistent with the standard feature set used in prior sepsis reinforcement learning studies, facilitating comparability with existing work \cite{komorowski2018artificial,raghu2017continuous,wu2023value,zhang2024optimizing,lu2024reinforcement}.

The action space $\mathcal{A}$ comprises discretized treatment decisions for intravenous (IV) fluids and vasopressors. Each intervention is discretized into five levels based on empirical quantiles, yielding $5 \times 5 = 25$ joint actions.

{\renewcommand{\arraystretch}{0.6} 

\begin{table}[H]
\caption{State Space Features}
\label{tab:included_variables}
\centering
\footnotesize
\begin{threeparttable}
\setlength{\tabcolsep}{3pt} 
\begin{tabular}{p{2.8cm}p{6.5cm}}
\toprule
\textbf{Category} & \textbf{Variables} \\
\midrule
Demographics/Scores (8) & Age, Gender, Weight, ICU readmission, GCS, Elixhauser s, SOFA, SIRS \\
Vitals (11) & HR, SBP, MBP, DBP, Resp. rate, Temperature, PaCO\textsubscript{2}, PaO\textsubscript{2}, PF ratio, SpO\textsubscript{2}, Shock index \\
Labs (24) & Albumin, pH, Calcium, Glucose, Hemoglobin, Magnesium, WBC, Creatinine, Bicarbonate, Sodium, Lactate, Chloride, Platelets, Potassium, PTT, PT, AST, ALT, BUN, INR, Ionised calcium, Total bilirubin, Base excess, Phosphate \\
Treatments and Interventions (2) & Mechanical ventilation, FiO\textsubscript{2} \\
Others (3) & Urine output (4-hourly), Total output, Time since sepsis start \\
\bottomrule
\end{tabular}

\vspace{0.5em}
\raggedright
\textbf{Abbreviations:} DBP = diastolic blood pressure;  
GCS = Glasgow Coma Scale; HR = heart rate; ICU = intensive care unit;  
MBP = mean blood pressure; PTT = partial thromboplastin time;  
Resp. rate = respiratory rate; SBP = systolic blood pressure; WBC = white blood cell count.
\end{threeparttable}
\end{table}

\subsection{\textbf{Preference-Based Reward Learning}}

Using narrative-derived TQS, we construct pairwise trajectory preferences and learn a reward function over state--action pairs.

Given trajectories $\mathcal{D} = \{\tau_i\}_{i=1}^N$ with scores $y_i$, we define:
\begin{equation}
\tau_i \succ \tau_j \quad \text{if} \quad y_i > y_j.
\end{equation}

Preference pairs are constructed by sampling trajectories against lower-scored trajectories, with constraints on minimum score differences and total pair count for stability.

Trajectory preferences are modeled using a Bradley--Terry formulation \cite{hunter2004mm}:
\begin{equation}
P_\theta(\tau_i \succ \tau_j) =
\frac{\exp\big(R_\theta(\tau_i)\big)}{\exp\big(R_\theta(\tau_i)\big) + \exp\big(R_\theta(\tau_j)\big)},
\end{equation}
where the trajectory return is defined as:
\begin{equation}
R_\theta(\tau_i) = \sum_{t=1}^{T_i} r_\theta(s_t, a_t).
\end{equation}

\paragraph{Weighted Margin-Based Objective}

For each pair $(\tau_i, \tau_j)$, we define the score gap:
\begin{equation}
\Delta_{ij} = y_i - y_j, \quad \text{with } \Delta_{ij} > 0,
\end{equation}
and task relevance scores $\eta_i, \eta_j \in [0,1]$, where higher values indicate greater relevance to sepsis-related decision-making.

We optimize:
\begin{equation}
\mathcal{L} = 
\mathbb{E}_{(i,j) \sim \mathcal{P}} \left[
w_{ij} \cdot 
\log\left(1 + \exp\left(m_{ij} - \big(R_\theta(\tau_i) - R_\theta(\tau_j)\big)\right)\right)
\right]
+ \lambda \, \mathbb{E}_{(s,a) \sim \mathcal{D}} \left[r_\theta(s,a)^2\right],
\end{equation}
with:
\begin{align}
w_{ij} &= \Delta_{ij} \, \eta_i \eta_j, \\
m_{ij} &= m_0 + \alpha \, \Delta_{ij}.
\end{align}

The weighting term $w_{ij}$ increases the contribution of trajectory pairs with larger score differences and higher task relevance, while the adaptive margin $m_{ij}$ enforces stronger separation between trajectories with greater outcome differences. The regularization term $\lambda$ controls reward magnitude.

Optimization was performed using Adam with gradient clipping and early stopping. The learned reward can be interpreted as a continuous proxy for trajectory-level clinical quality.

\paragraph{Reward Model Architecture}

The reward function is parameterized by a neural network over state--action pairs. Discretized IV fluid and vasopressor actions are represented using separate learnable embeddings (5 bins each, 8-dimensional):
\[
\mathbf{e}^{\text{IV}}(a_t^{\text{IV}}) \in \mathbb{R}^8, 
\quad
\mathbf{e}^{\text{vaso}}(a_t^{\text{vaso}}) \in \mathbb{R}^8.
\]

These embeddings are concatenated with the state vector to form the input:
\[
\mathbf{x}_t = \big[\, \mathbf{s}_t, \, \mathbf{e}^{\text{IV}}(a_t^{\text{IV}}), \, \mathbf{e}^{\text{vaso}}(a_t^{\text{vaso}}) \,\big].
\]

The combined representation is passed through a two-layer multilayer perceptron (hidden size 128) with ReLU activation and dropout ($p = 0.2$). The network outputs a scalar reward for each time step, which is subsequently post-processed to improve numerical stability and suitability for downstream policy learning.

\paragraph{Reward Normalization}

To improve numerical stability, per-step rewards were clipped at the 1st and 99th percentiles and subsequently scaled using a hyperbolic tangent transformation:
\begin{equation}
r_t^{\text{final}} = \tanh\left(\frac{\tilde{r}_t}{c}\right),
\end{equation}
where $\tilde{r}_t$ denotes the raw reward output by the network. This transformation bounds rewards to $[-1,1]$ while preserving relative ordering.

\subsection{\textbf{Policy Learning (Offline RL)}}

Patient trajectories were randomly split into training (80\%) and testing (20\%) sets. To isolate the effect of reward formulation, all policies were trained using the same offline RL algorithm (Dueling Double DQN with Conservative Q-Learning; D3QN-CQL) with identical architectures and hyperparameters. This controlled setup ensures that differences in policy behavior are attributable to reward formulation rather than optimization dynamics.

The Q-network architecture consisted of two fully connected layers (hidden size 128) with Leaky ReLU activation. Models were trained using Adam with learning rate $10^{-3}$, discount factor $\gamma = 0.99$, batch size 256, and CQL regularization parameter $\alpha = 0.5$, following established implementations used in prior clinical RL studies \cite{raghu2017continuous,roggeveen2021transatlantic,wu2023value,tan2024advancing}.

\subsection{\textbf{Baselines}}

We compare CN-PR against several commonly used reward formulations in RL-based dynamic treatment regimes. These baselines represent common reward design strategies in the literature, spanning both outcome-based and handcrafted physiological formulations.

\paragraph{Terminal Mortality Reward}

A widely used baseline \cite{komorowski2018artificial,roggeveen2021transatlantic,luo2024reinforcement,tang2022leveraging,killian2020empirical} defines reward solely based on terminal survival outcome. Specifically, rewards are zero at all intermediate time steps, with a terminal reward of $+R$ assigned to survivors and $-R$ assigned to non-survivors:
\begin{equation}
r_t =
\begin{cases}
0, & t < T \\
+R, & t = T \text{ and patient survives} \\
-R, & t = T \text{ and patient dies}
\end{cases}
\end{equation}
where $R$ is a positive constant. This formulation directly optimizes survival but provides no intermediate feedback, resulting in sparse rewards and challenging temporal credit assignment.

\paragraph{SOFA-Lactate Reward (SOFA-Lac)}

To incorporate intermediate clinical signals, we consider a reward based on organ failure severity and metabolic dysfunction. Let $\text{SOFA}_t$ denote the Sequential Organ Failure Assessment score and $v_t$ denote lactate level at time $t$. Following prior work \cite{raghu2017continuous,wu2023value,zhang2024optimizing,tan2024advancing}, the reward integrates both absolute levels and temporal changes:

\begin{equation}
\begin{aligned}
r_t = \;&
c_0 \cdot \tanh\!\left(\sum_{k} \mathbf{1}_{\text{SOFA}_{t,k} > 0}\right)
+ c_1 \cdot (\text{SOFA}_{t+1} - \text{SOFA}_t) \\
&+ c_2 \cdot \tanh(v_{t+1} - v_t)
+ \mathbf{1}_{t=T} \cdot r_{\text{outcome}}.
\end{aligned}
\end{equation}

where $c_0, c_1, c_2$ are fixed coefficients, $\text{SOFA}_{t,k}$ denotes organ-specific SOFA components, and $r_{\text{outcome}}$ is a terminal reward reflecting survival. This formulation penalizes both persistent and worsening organ dysfunction while incorporating metabolic instability through lactate dynamics.

\paragraph{NEWS2-Based Reward}

We also evaluate a reward based on the National Early Warning Score 2 (NEWS2), which aggregates vital sign abnormalities into a composite clinical risk score \cite{luo2024reinforcement,inada2018news}. NEWS2 was normalized to the range $[0,1]$ to ensure comparability across patients, with higher values indicating greater physiological risk. The reward is defined as:
\begin{equation}
r_t = -\text{NEWS2}_t + \mathbf{1}_{t=T} \cdot r_{\text{outcome}},
\end{equation}
where $r_{\text{outcome}}$ encodes terminal survival outcome. This formulation provides a dense proxy for physiological deterioration while maintaining alignment with mortality. A full specification of the NEWS2 reward function is provided in Supplementary Information Section~1.2.

\paragraph{Random Policy}

As a sanity check, we include a random policy baseline that selects treatment actions uniformly at random from the discrete action space:
\[
\pi_r(a) = \frac{1}{|\mathcal{A}|},
\]
where $|\mathcal{A}|$ denotes the total number of possible actions.

This baseline serves as a lower bound for policy performance and helps verify that observed policy--outcome associations are not attributable to generic action variability alone.

Together, these baselines capture a range of commonly used reward design strategies in clinical RL, spanning sparse outcome-based objectives and manually engineered physiological rewards.

\subsection{\textbf{Policy Evaluation}}
\label{sec:policy_eval}

Finally, we evaluate learned policies using complementary outcome-based and behavioral analyses. In addition to assessing associations with clinical outcomes, we characterize the structure of learned treatment strategies and the clinical factors influencing decision-making. This multi-faceted evaluation provides a more comprehensive assessment of whether learned policies are both clinically aligned and behaviorally plausible.

\subsubsection{Regression-Based Evaluation}

We assess the relationship between policy--clinician disagreement and clinical outcomes using multivariable regression.

For each trajectory $\tau_i$, we compare the actions recommended by policy $\pi$ with those taken by clinicians at each time step. Let $a_{t,i}^{\pi}$ denote the policy action and $a_{t,i}^{\text{clin}}$ the observed clinician action, where each action is a two-dimensional vector:
\[
a_{t,i} = \big(a_{t,i}^{\text{IV}}, \, a_{t,i}^{\text{vaso}}\big).
\]

We compute per-step absolute differences:
\begin{align}
d_{t,i}^{\text{IV}} &= \left| a_{t,i}^{\text{IV},\pi} - a_{t,i}^{\text{IV},\text{clin}} \right|, \\
d_{t,i}^{\text{vaso}} &= \left| a_{t,i}^{\text{vaso},\pi} - a_{t,i}^{\text{vaso},\text{clin}} \right|,
\end{align}
and define the joint Euclidean distance:
\begin{equation}
d_{t,i}^{\text{joint}} =
\sqrt{
\left(d_{t,i}^{\text{IV}}\right)^2 +
\left(d_{t,i}^{\text{vaso}}\right)^2
}.
\end{equation}

Distances are aggregated at the trajectory level:
\begin{equation}
\bar{d}_i^{\text{joint}} =
\frac{1}{T_i}
\sum_{t=1}^{T_i} d_{t,i}^{\text{joint}},
\end{equation}
and standardized (z-scored) prior to regression.

For each policy $\pi$, we fit:
\begin{equation}
Y_i =
\beta_0 +
\beta_1 \bar{d}_i^{\text{joint}} +
\boldsymbol{\beta}^\top \mathbf{X}_i +
\epsilon_i,
\end{equation}
where $Y_i$ denotes the outcome for patient $i$, $\bar{d}_i^{\text{joint}}$ is the average joint distance, and $\mathbf{X}_i$ represents baseline covariates.

We evaluate policies across complementary outcome measures capturing clinically relevant dimensions of recovery beyond mortality alone. Collectively, these outcomes assess early physiologic recovery, disease trajectory, functional outcome, and treatment intensity, whereas mortality alone provides only a coarse summary of patient outcome. The evaluated outcomes are:

\begin{itemize}

    \item \textbf{Hospital mortality:} binary indicator of in-hospital death (1 = death, 0 = survival).

    \item \textbf{Organ support--free days (OSFD-7):} days alive and free from vasopressors, mechanical ventilation, and renal replacement therapy within the first 7 days \cite{russell2018days}.

    \item \textbf{Time to shock resolution (TSR):} time from sepsis onset to sustained hemodynamic stability without vasopressor support.

    \item \textbf{Discharge disposition score:} proxy for functional recovery based on discharge status.

    \item \textbf{Treatment burden:} mean IV fluid and vasopressor administration per time step, reflecting treatment intensity independent of trajectory length.

\end{itemize}

Full definitions are provided in Supplementary Information Section~1.3.

Covariates include age, baseline SOFA score, Elixhauser comorbidity index, lactate, shock index, and mechanical ventilation status. Continuous outcomes are modeled using linear regression with heteroskedasticity-robust (HC3) standard errors, while binary outcomes are modeled using logistic regression.

The coefficient $\beta_1$ quantifies the association between policy--clinician disagreement and clinical outcomes. Greater agreement with a policy is associated with improved trajectories when deviation is negatively associated with favorable outcomes (e.g., OSFD) or positively associated with adverse outcomes (e.g., treatment burden). All analyses are performed at the patient level.

This regression-based framework was selected because our primary objective is to compare reward formulations rather than benchmark RL algorithms. Standard off-policy evaluation (OPE) estimates policy value under a fixed reward, whereas here each policy is trained using a different reward definition, making direct cross-reward comparisons ill-posed. Instead, we evaluate whether learned policies align with clinically meaningful outcomes independent of the training objective, thereby mitigating reward circularity. This approach is consistent with prior work comparing reward formulations in healthcare RL \cite{lu2024reinforcement,lin2018deep}. Importantly, these analyses quantify associations rather than causal effects and should therefore be interpreted as evidence of clinical alignment rather than policy optimality.

\subsubsection{Action Distribution Analysis}

To characterize the treatment strategies learned by each policy, we analyze the empirical distribution of actions across the joint IV fluid--vasopressor space. For each policy, we compute the frequency of each discretized action pair and visualize these distributions using two-dimensional heatmaps.

To assess whether policies adapt to patient severity, action distributions are stratified by baseline SOFA score into low- and high-severity groups. This enables evaluation of whether policies exhibit clinically plausible escalation or de-escalation patterns across severity strata.

\subsubsection{Feature Importance Analysis}

To further understand the clinical factors driving treatment decisions, we perform feature importance analysis using supervised models that approximate each policy. For each intervention (IV fluids and vasopressors), we train separate random forest classifiers to predict the discrete action selected at each time step under either the clinician policy or the CN-PR policy.

Feature importance is quantified using permutation importance, which measures the change in predictive performance when individual features are randomly permuted. This provides a model-agnostic estimate of the relative contribution of each variable to treatment decision-making.

\subsection{\textbf{External Validation}}

To assess generalizability, we conducted external validation using the AmsterdamUMCdb dataset, an open-access critical care database containing ICU admissions from Amsterdam University Medical Center between 2003 and 2016 \cite{thoral2021sharing}. Using the same Sepsis-3 inclusion criteria and preprocessing pipeline as the primary MIMIC-IV analysis, we extracted a cohort of 4{,}404 patients. Full cohort characteristics are provided in Supplementary Table~S1.

Because AmsterdamUMCdb does not contain several variables used in the primary analysis (PT, INR, CO\textsubscript{2}, and Elixhauser comorbidity score), we adopted a reduced-feature strategy using only variables shared across both datasets. A secondary CN-PR reward model was therefore trained within this harmonized feature space and directly transferred to AmsterdamUMCdb without further fine-tuning. Reward generalization was evaluated using correlations between trajectory-level rewards and clinically relevant outcomes, including mortality, OSFD-7, TSR, and treatment burden. Reinforcement learning policies were subsequently re-trained and evaluated on AmsterdamUMCdb using the same offline RL and regression-based evaluation framework as in the primary analysis. Because discharge disposition and Elixhauser comorbidity scores were unavailable in AmsterdamUMCdb, discharge score and Elixhauser-adjustment were excluded from external validation analyses.


\section{Results}

\subsection{Validation of the Learned Reward Function}

\subsubsection{Distribution of Narrative-Derived TQS}

Figure~\ref{fig:score_distribution} illustrates the distribution of TQS derived from clinical narratives. Most trajectories received scores of 4--5, corresponding to survival with varying degrees of recovery, whereas lower scores captured mortality and severe morbidity. The presence of intermediate categories indicates that narrative-derived labeling provides a graded assessment of clinical trajectory quality beyond binary outcomes.

Task relevance scores remained relatively stable across TQS categories, suggesting that task relevance captured narrative informativeness for sepsis management rather than merely reflecting overall outcome severity.

Although the cohort exhibited a right-skewed distribution toward favorable outcomes, preference learning relies on relative comparisons rather than class balance. Moreover, the proposed objective emphasizes larger score differences through score-gap weighting and margin-based separation, thereby prioritizing clinically meaningful distinctions between trajectories.

\begin{figure}[!hbt]
\centering
\includegraphics[width=0.75\columnwidth]{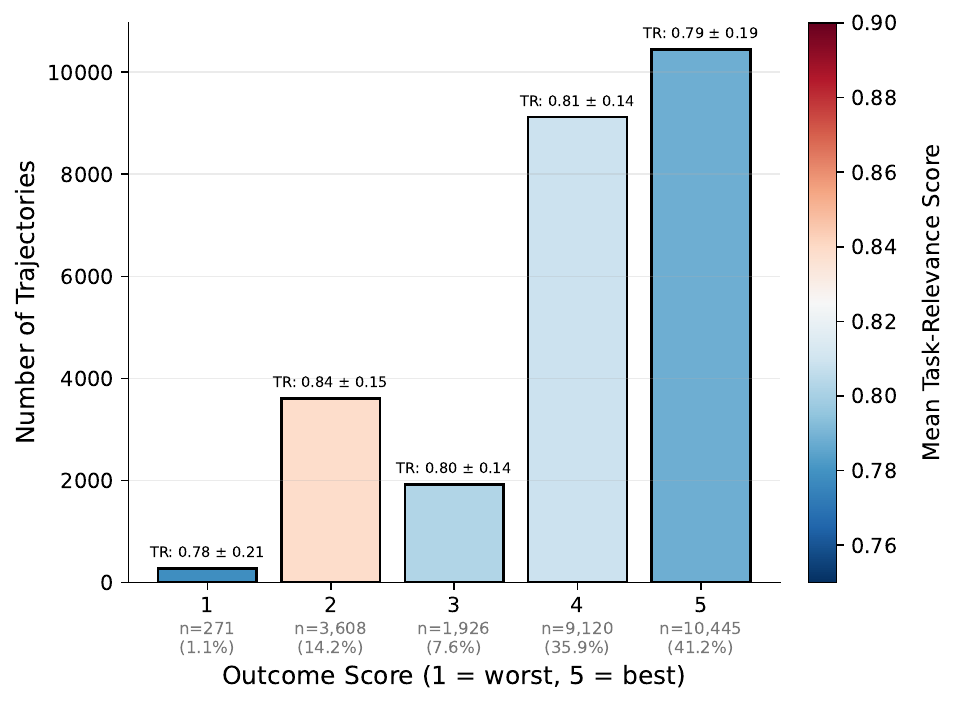}
\caption{
Distribution of TQS (1--5) derived from clinical narratives across the full study cohort. Bars indicate the number of trajectories per score category, with percentages shown below each bar. Bar color encodes the mean task relevance score within each category.
}
\label{fig:score_distribution}
\vspace{-0.5em}
\end{figure}


\subsubsection{Alignment with Trajectory Quality Scores}

Agreement between the LLM and clinicians for TQS was comparable to inter-clinician agreement (Supplementary Table~S3), with high quadratic-weighted Cohen’s $\kappa$ ($0.910$ vs. $0.917$), strong Spearman correlations ($0.923$ vs. $0.931$), and near-perfect within-one-category agreement ($98.9\%$ vs. $99.8\%$). Clinician assessment of task relevance similarly demonstrated substantial agreement with LLM-derived task relevance scores, supporting the reliability of the proposed task-aware weighting mechanism.

Figure~\ref{fig:reward_boxplot} shows that the learned reward increases monotonically with TQS, with clear separation between low- and high-quality trajectories. This ordering was supported by a strong monotonic association (Spearman $\rho = 0.63$, $p < 0.001$) and a large effect size between extreme groups (Cohen’s $d = 3.41$, 95\% CI: [3.03, 3.82]), indicating that the reward reliably recovered the underlying preference structure.

\begin{figure}[H]
\centering
\includegraphics[width=0.75\columnwidth]{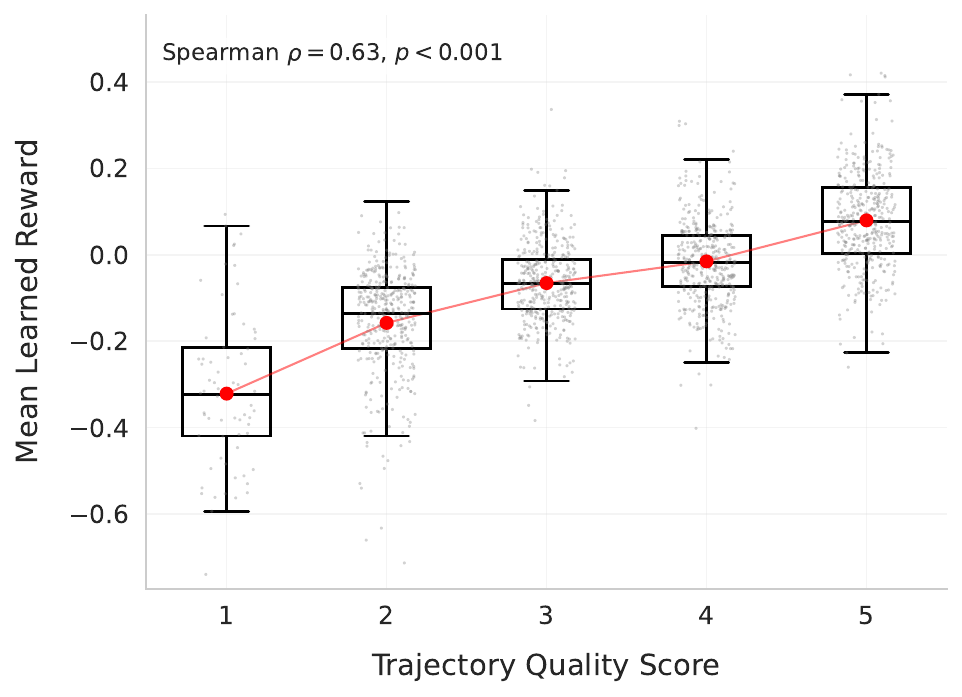}
\caption{
Per-trajectory mean learned reward stratified by TQS (1 = lowest, 5 = highest) on the test set. Red markers denote group means.
}
\label{fig:reward_boxplot}
\vspace{-0.5em}
\end{figure}

Correlation analyses with downstream clinical outcomes (Table~\ref{tab:reward_correlations_mimic}) further demonstrated that the learned reward aligned with key endpoints, including mortality, organ support--free days, and treatment burden. Notably, the CN-PR reward exhibited stronger and more consistent associations with these outcomes than alternative handcrafted and outcome-based reward formulations, and these patterns remained consistent in the external AmsterdamUMCdb validation cohort (Supplementary Table~S4). Ablation of the task-relevance weighting component resulted in attenuated reward--outcome correlations across both internal and external validation datasets, particularly for organ support--free days, time to shock resolution, and treatment burden metrics. These findings suggest that task-aware supervision improves alignment between narrative-derived preferences and clinically relevant sepsis management outcomes.

\begin{table}[h]
\centering
\caption{
Spearman correlation between trajectory-level cumulative rewards and clinical outcomes for different reward formulations on the internal MIMIC-IV dataset.
}
\label{tab:reward_correlations_mimic}
\resizebox{\columnwidth}{!}{
\begin{tabular}{lcccccc}
\toprule
\textbf{Reward} & \textbf{OSFD-7} & \textbf{TSR} & \textbf{Mortality} & \textbf{IV Fluid Burden} & \textbf{Vasopressor Burden} & \textbf{Discharge Score} \\
\midrule

CN-PR 
& \textbf{0.48} 
& \textbf{-0.22} 
& -0.57 
& \textbf{-0.18} 
& \textbf{-0.30} 
& \textbf{0.59} \\

CN-PR (No TR) 
& 0.44 
& -0.19 
& -0.55 
& -0.16 
& -0.26 
& 0.55 \\

Mortality-only 
& 0.38 
& -0.08 
& \textbf{-0.64} 
& -0.02 
& -0.15 
& 0.41 \\

SOFA-Lac 
& 0.30 
& -0.03 
& -0.63 
& -0.06 
& -0.08 
& 0.40 \\

NEWS2 
& 0.22 
& -0.12 
& -0.48 
& -0.07 
& -0.17 
& 0.41 \\

Permuted-TQS 
& -0.06 
& 0.10 
& 0.03 
& -0.02 
& 0.12 
& -0.02 \\

\bottomrule
\end{tabular}
}
\vspace{0.3em}

\footnotesize
\textit{Notes:} OSFD-7 = organ support--free days at 7 days; TSR = time to shock resolution; TR = task-relevance weighting. 
\end{table}

As a negative control, we repeated reward learning after randomly permuting TQS labels across trajectories while preserving the empirical score distribution. The resulting rewards showed weak and inconsistent correlations with clinical outcomes (all $|\rho| < 0.12$; Table~\ref{tab:reward_correlations_mimic}), indicating that the observed alignment did not arise from spurious statistical structure or the reward-learning procedure alone.


\subsubsection{Counterfactual Treatment Sensitivity}

Figure~\ref{fig:reward_heatmaps} illustrates counterfactual reward surfaces across severity strata. In lower-severity patients (SOFA $<5$), higher fluid and vasopressor dosing was consistently penalized, whereas in higher-severity patients (SOFA $>15$), moderate treatment levels were favored relative to minimal intervention.

Across severity strata, the reward exhibited smooth non-linear structure with well-defined regions of higher expected return, reflecting clinically plausible trade-offs between treatment benefit and intervention intensity. Unlike sparse outcome-based rewards, CN-PR provides dense and temporally informative feedback throughout the clinical trajectory.

\begin{figure*}[h]
\centering
\includegraphics[width=1\textwidth]{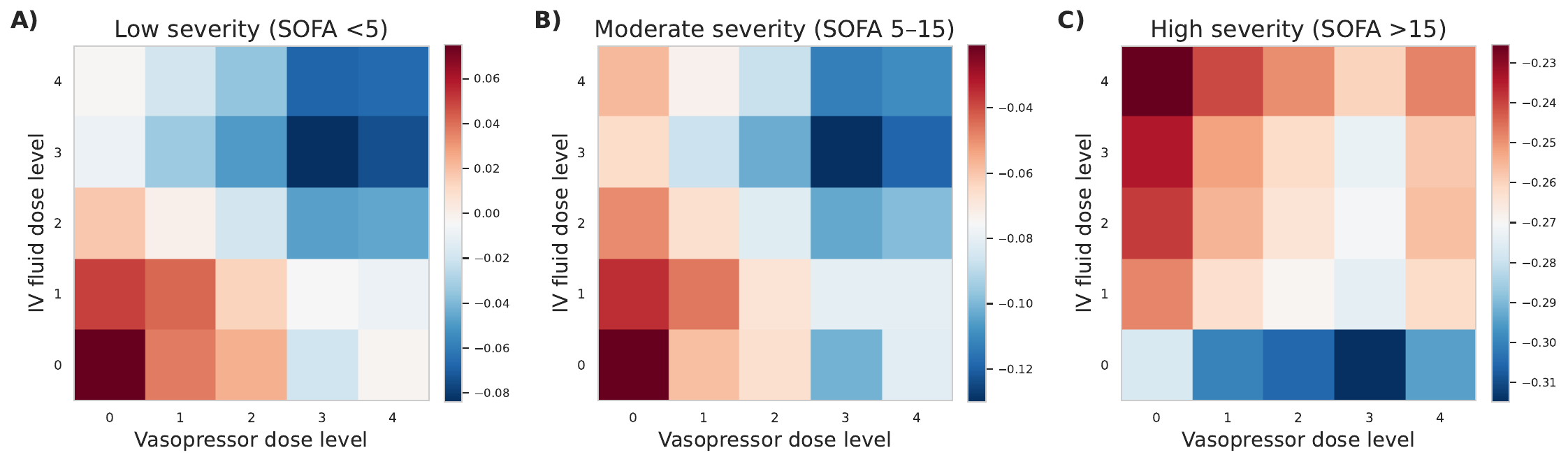}
\caption{
Counterfactual joint treatment reward surfaces across severity strata. Each panel shows the expected reward as a function of IV fluid and vasopressor dosing levels for different patient groups: (A) low severity (SOFA < 5), (B) moderate severity (SOFA 5--15), and (C) high severity (SOFA > 15). Heatmap values represent the predicted reward for each joint treatment combination, illustrating how optimal dosing strategies vary with patient severity.
}
\label{fig:reward_heatmaps}
\vspace{-0.5em}
\end{figure*}

\FloatBarrier

\subsection{Policy Evaluation}

Across outcomes, the CN-PR policy achieved performance comparable to outcome-based baselines for mortality while exhibiting stronger associations with recovery-related metrics and treatment burden (Figure~\ref{fig:policy_discrepancy}).

For mortality, the CN-PR policy showed a strong positive association between policy deviation and risk ($\beta = 0.48$), closely matching mortality- and SOFA-Lac--based policies (both $\beta \approx 0.50$), indicating comparable alignment with survival outcomes.

In contrast, CN-PR demonstrated the largest effect sizes for recovery-oriented outcomes. The policy exhibited the strongest associations for OSFD-7 ($\beta = -0.84$) and TSR ($\beta = 7.79$), indicating that agreement with the policy was more strongly associated with improved recovery trajectories relative to baseline reward formulations. Importantly, these gains were not attributable to reward density alone. Although both CN-PR and NEWS2-based rewards provide dense supervision, CN-PR consistently demonstrated stronger associations with clinically relevant outcomes, suggesting that the quality and clinical relevance of the learned reward signal are critical determinants of downstream policy behavior. 

\begin{figure}[!t]
\centering
\includegraphics[width=1\textwidth]{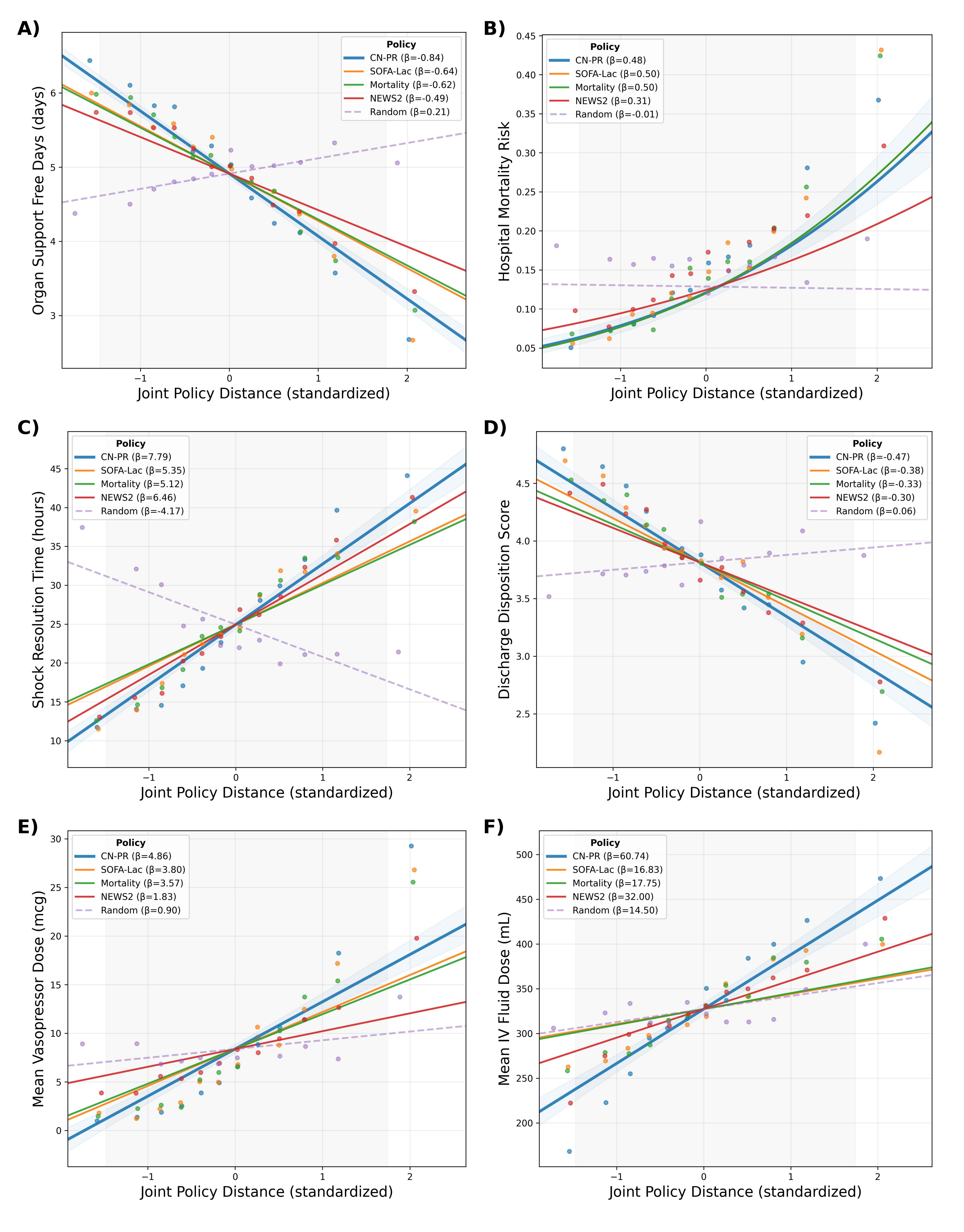}
\caption{
Relationship between policy--clinician discrepancy (measured as joint action distance) and clinical outcomes across multiple endpoints. Each subplot shows model-adjusted outcomes as a function of standardized joint action distance: A) OSFD-7, B) hospital mortality risk, C) TSR, D) discharge disposition score, E) mean vasopressor dose, and F) mean IV fluid dose. Curves represent regression-adjusted estimates for each policy. Shaded confidence intervals are shown for the CN-PR policy only to improve visual clarity; full regression results are provided in Supplementary Table~S5.
}
\label{fig:policy_discrepancy}
\vspace{-0.5em}
\end{figure}

\FloatBarrier

The largest differences were observed for treatment burden. Deviations from the CN-PR policy corresponded to substantially larger increases in both vasopressor and IV fluid administration, reflected by steeper regression slopes relative to baseline policies. These findings suggest that the CN-PR policy achieves favorable outcomes with lower intervention intensity, whereas outcome-based rewards are associated with comparatively more aggressive treatment strategies.

These patterns were preserved during external validation on the AmsterdamUMCdb cohort. Although absolute effect sizes were attenuated, the CN-PR policy maintained comparable performance for mortality while retaining stronger associations with recovery-related outcomes and treatment burden relative to baseline rewards (Supplementary Table~S6). The relative ordering of policies remained consistent across datasets, supporting the robustness and generalizability of the learned reward.

By contrast, the random policy demonstrated weak or inconsistent associations across all outcomes in both datasets, indicating that the observed relationships were not driven by generic action variability alone. Although these analyses are observational rather than causal, the findings suggest that alignment with the CN-PR policy is associated with more favorable and treatment-efficient clinical trajectories.

\paragraph{Policy Behavior and Clinical Alignment}

Action distribution heatmaps (Figure~\ref{fig:action_heatmaps}) further illustrate these differences. Clinician actions were concentrated in low-intensity treatments across severity strata, whereas the CN-PR policy exhibited more differentiated and severity-adaptive distributions. For lower-severity patients, the policy favored minimal intervention, while for higher-severity patients it shifted toward moderate fluid administration with selective vasopressor use, reflecting clinically plausible escalation of care.

\begin{figure}[h]
\centering
\includegraphics[width=1\textwidth]{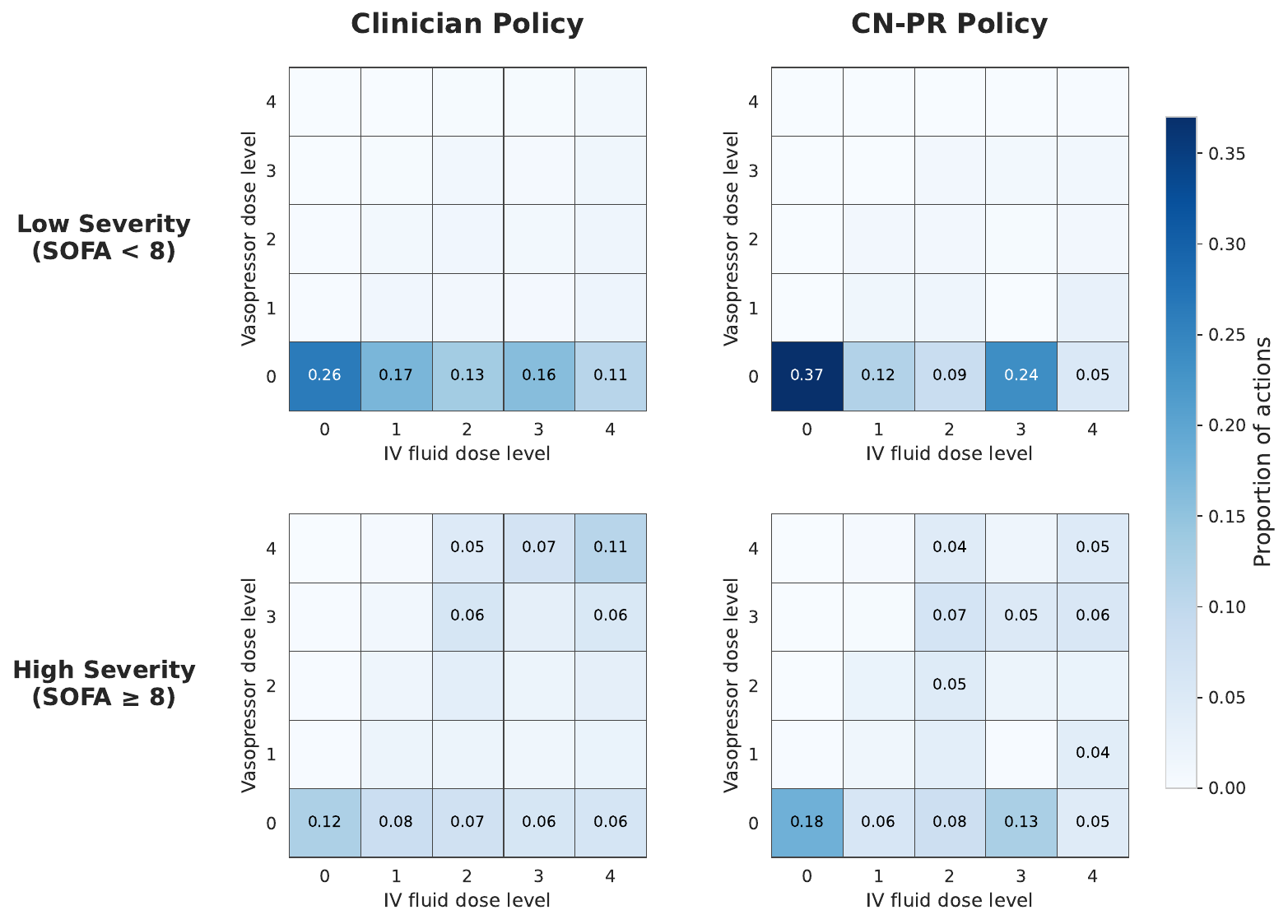}
\caption{
Joint action distributions for IV fluids and vasopressors under clinician and CN-PR policies, stratified by severity (Low: SOFA $<8$; High: SOFA $\geq 8$). Cells show action proportions; only values $\geq 0.04$ are annotated.
}
\label{fig:action_heatmaps}
\vspace{-0.5em}
\end{figure}

Feature importance analyses (Supplementary Figure~S1) showed that both clinician and CN-PR policies were primarily driven by core clinical variables, including urine output, SOFA score, and temporal features such as time since sepsis onset. These findings support the clinical plausibility of the learned policy and suggest that treatment decisions are guided by meaningful physiological signals rather than spurious correlations.


\section{Discussion and Conclusion}

In this work, we introduced CN-PR, a framework for learning reward functions from clinical narratives by leveraging discharge summaries as implicit evaluations of patient trajectories. Rather than relying on manually specified objectives or handcrafted physiologic proxies, CN-PR reframes reward specification as a problem of extracting preference signals from routinely collected clinical documentation.

Policies trained using CN-PR achieved performance comparable to outcome-based baselines for mortality while demonstrating stronger associations with recovery-oriented outcomes and treatment burden. Importantly, this pattern is clinically plausible and reflects the multidimensional nature of critical care outcomes. Prior work in septic shock has shown that clinically meaningful differences in organ dysfunction and organ-support requirements may exist even in the absence of detectable mortality differences, particularly as overall sepsis mortality declines \cite{russell2018days,linder2016short}. Recovery-oriented endpoints such as organ support–free days and treatment burden therefore capture complementary dimensions of patient recovery that are not fully reflected by hospital survival alone. Mortality is a coarse endpoint that does not distinguish between heterogeneous recovery trajectories, duration of organ support, treatment intensity, or overall recovery efficiency. In contrast, narrative-derived supervision may capture broader dimensions of clinical recovery, enabling policies that preserve survival performance while more closely aligning with recovery-oriented outcomes and treatment burden. Consistent with prior literature linking reduced organ dysfunction with improved longer-term outcomes \cite{linder2016short}, these findings suggest that narrative-informed reward learning may better capture clinically meaningful differences in recovery quality and intervention efficiency beyond survival alone \cite{gavelli2021management,lesur2018hemodynamic}. Comparisons with alternative dense reward formulations further suggest that these improvements are not attributable to reward density alone, but instead reflect the ability of narrative-derived supervision to capture clinically meaningful structure within patient trajectories.

These findings highlight an important limitation of commonly used reward formulations in healthcare RL. Outcome-based and handcrafted rewards are typically derived from a narrow set of predefined endpoints and therefore capture only a partial representation of clinical success. In contrast, clinical narratives implicitly encode broader aspects of care, including treatment response, reversibility, physiologic stability, and recovery trajectory. By learning from these narrative-derived assessments, CN-PR provides a more expressive representation of patient outcome quality without requiring explicit specification of intermediate objectives.

More broadly, our results reinforce the importance of reward specification as a primary determinant of policy behavior in offline RL for healthcare \cite{luo2024reinforcement,liang2025methods,lu2024reinforcement}. Rather than constructing rewards through heuristic combinations of clinical variables, CN-PR enables reward learning directly from data sources that encode high-level clinical reasoning. The proposed task-relevance weighting mechanism further improves robustness by reducing the influence of weakly informative or noisy supervision signals, acknowledging that not all clinical narratives are equally informative for downstream treatment optimization.

The framework is also compatible with clinician-in-the-loop decision support systems \cite{topol2019high,sendak2020real}. Because policy inputs rely on routinely collected structured data, integration into existing clinical workflows is feasible, while narrative-derived reward learning remains an offline process. This separation allows reward functions to reflect clinician reasoning without requiring real-time access to unstructured text during deployment. Prospective validation, including silent deployment and simulation-based evaluation, will nevertheless be necessary to assess safety, usability, and clinical impact \cite{yu2021reinforcement,zhou2021clinical}.

Several limitations warrant consideration. First, the learned reward depends on LLM-based interpretation of clinical narratives and may inherit biases present in either documentation practices or model inference. Variability in note quality may also affect the reliability of extracted signals, although the proposed task-relevance weighting mechanism partially mitigates this issue. Nevertheless, clinician validation demonstrated strong agreement between LLM-derived and clinician-assigned trajectory quality and task relevance scores, supporting the clinical consistency of the extracted supervision signals. Second, cohort construction required temporally aligned discharge summaries, potentially introducing selection bias toward more complete and well-documented trajectories. Finally, our evaluation framework is observational and does not establish causal effects, leaving the possibility of residual confounding despite adjustment for key covariates \cite{liang2025methods,yu2021reinforcement}.

Future work could extend this framework by incorporating longitudinal clinical notes to learn time-varying reward signals and by explicitly modeling uncertainty in narrative-derived supervision. Integrating preference-based reward learning with causal inference techniques may further improve robustness and support safer policy evaluation.

Beyond critical care, the proposed framework may generalize to domains where narrative documentation encodes trajectory-level evaluation, including oncology, chronic disease management, and psychiatry, where clinical notes frequently capture treatment response, progression, functional status, and tolerability over time \cite{yu2021reinforcement,kehl2020natural,yang2023reinforcement,tejedor2020reinforcement,barak2017predicting}.

In summary, CN-PR demonstrates how routinely collected clinical narratives can be leveraged as scalable supervision for reward learning in dynamic treatment regimes, enabling more expressive and clinically grounded reward design for healthcare RL.

\section{Ethical Statement}

Consistent with prior work on reinforcement learning–based clinical decision support systems for sepsis and critical care \cite{komorowski2018artificial,raghu2017continuous,wu2023value}, our framework is intended to support rather than replace clinician decision-making. The system provides data-driven recommendations derived from historical clinical trajectories, while treatment decisions remain under clinician oversight and responsibility. This human-in-the-loop design preserves clinical autonomy and enables integration of machine-assisted recommendations with expert medical judgment.

All models were developed and evaluated exclusively using de-identified, publicly available retrospective datasets, without real-time clinical deployment or patient interaction. Consequently, the study posed no direct risk to patients and was conducted in accordance with standard ethical practices for secondary analysis of anonymized health data.

\section{Data Availability}
The structured open-source MIMIC-IV 2.0 data used in this study is available at 
\url{https://physionet.org/content/mimiciv/2.0/}. The clinical notes data is available at \url{https://www.physionet.org/content/mimic-iv-note/2.2/}.

\section{Code Availability}
The underlying code for this study is available on Github via \url{https://github.com/danjst/CNPR}.

\section{Declaration of generative AI and AI-assisted technologies in the manuscript preparation process
}
During the preparation of this work, the authors used ChatGPT (OpenAI, San Francisco, CA, USA) to assist with language refinement and grammatical editing. After using this tool, the authors reviewed and edited the content as needed and take full responsibility for the content of the published article.

\bibliographystyle{elsarticle-num}
\bibliography{references}

\section{Funding}

This research was supported by A*STAR, CISCO Systems (USA) Pte. Ltd., and the National University of Singapore under the Cisco–NUS Accelerated Digital Economy Corporate Laboratory (Award I21001E0002); the National University of Singapore President’s Graduate Fellowship; the AI for Public Health Program at the Saw Swee Hock School of Public Health, National University of Singapore; and the NUHS Innovation Grant (NUHSRO/2025/088/RO5+5/Innovation/LOA03).

\section{Author Contributions}
MF is the guarantor of the content of the manuscript, including the data and analysis. DJT conceptualized the study, designed the algorithm, collected data and results, and was the primary manuscript writer. KCS contributed to data analysis and interpretation. JHZC, KWH, and AYYN performed clinical validation of the LLM-derived outputs. All authors participated in manuscript review, provided final approval of the manuscript, and take responsibility for the accuracy and integrity of the work. 

\section{Conflicts of Interest}
All authors declare no financial or non-financial conflicts of interest.

\clearpage

\setcounter{section}{0}
\setcounter{subsection}{0}
\setcounter{subsubsection}{0}

\renewcommand{\thesection}{\arabic{section}}
\renewcommand{\thesubsection}{\thesection.\arabic{subsection}}
\renewcommand{\thesubsubsection}{\thesubsection.\arabic{subsubsection}}

\vspace{0.5em}
\section*{\fontsize{16}{18}\selectfont\bfseries Supplementary Information}
\vspace{0.5em}

\addcontentsline{toc}{section}{Supplementary Information}

\renewcommand{\thefigure}{S\arabic{figure}}
\renewcommand{\thetable}{S\arabic{table}}
\setcounter{figure}{0}
\setcounter{table}{0}
\setlength{\tabcolsep}{4pt}     
\renewcommand{\arraystretch}{0.95}

\setlength{\tabcolsep}{4pt}
\renewcommand{\arraystretch}{0.95}

\captionsetup{justification=centering}
\captionsetup[table]{skip=4pt}

\makeatletter
\renewcommand{\section}{\@startsection{section}{1}{\z@}%
  {-2.5ex \@plus -1ex \@minus -.2ex}%
  {1.5ex \@plus .2ex}%
  {\normalfont\large\bfseries}}

\renewcommand{\subsection}{\@startsection{subsection}{2}{\z@}%
  {-2.0ex \@plus -1ex \@minus -.2ex}%
  {1.0ex \@plus .2ex}%
  {\normalfont\normalsize\itshape}}
\makeatother

\section{Supplementary Methods}

\subsection{LLM Prompt for Trajectory Quality Scoring}

The full prompt used to assign trajectory-level outcome scores from discharge summaries is provided below. The prompt was designed to emulate expert clinical reasoning, with explicit guidance on sepsis management, trajectory interpretation, and outcome categorization. Structured instructions were included to ensure consistency, enforce concise justification, and minimize ambiguity in scoring.

\begin{lstlisting}

[
You are an expert intensivist and clinical outcomes reviewer. You will read ICU/HDU/medicine discharge summaries and assign a single outcome score from 1-5 that reflects the overall quality and success of the ICU/hospital treatment course, with special attention to sepsis-related management (IV fluids, vasopressors, antibiotics, source control, organ support) and resulting outcomes.

The note is at the whole-hospital-stay level, but whenever possible you should pay particular attention to:
- ICU-level decisions (hemodynamics, vasopressors, fluid balance, ventilation, renal support, source control)
- Sepsis-related complications, shock, organ failures, and how well they were reversed or controlled
- The patient's final clinical trajectory and disposition

Use these scores:

1 = Deceased - catastrophic, abrupt, or early death with potentially modifiable elements
- Death occurs suddenly, early, or with a rapid collapse.
- Often occurs within the first 24-48 hours without a meaningful stabilization phase.
- Outcome appears out of proportion to presenting severity OR potentially salvageable components existed.
- Focus is on trajectory character (abrupt, catastrophic), not explicit preventability.

2 = Deceased - prolonged, expected, or biologically inevitable death due to overwhelming illness or baseline status
- Death follows a prolonged, predictable, or biologically inevitable decline.
- Strong evidence of:
  - Very poor baseline status (e.g., ESRD, metastatic cancer, frailty)
  - Multi-organ failure on arrival
  - Refractory shock/respiratory failure despite appropriate care
- Death reflects the natural history of severe disease rather than abrupt collapse.

3 = Survived, but with poor outcome / very high morbidity or near-terminal trajectory
- Patient survives this admission but has:
  - Major unresolved organ failure (e.g., NEW dialysis dependence, chronic ventilator/tracheostomy dependence, severe persistent heart failure)
  - Transition to hospice, CMO, DNR/DNI with non-escalation plan, or explicit near-terminal goals
  - Catastrophic irreversible neurologic or functional deficit
- Persistent NEW life-sustaining dependence at discharge strongly favors Score 3.
- Represents survival with permanent severe morbidity or an end-of-life trajectory.
- Discharge to an extended care facility alone does NOT imply Score 3.

4 = Survived, reasonable recovery but with significant ongoing morbidity
- Patient stabilizes and recovers, but with ongoing limitations.
- Organ failures were controlled or partially reversed.
- Recovery trajectory remains restorative.
- Prognosis is guarded but not near-terminal.

5 = Survived, good recovery / strong treatment success
- Patient survives with good functional recovery and minimal persistent organ dysfunction.
- Sepsis/shock management was timely and effective.
- Represents strong overall clinical success.

Scores represent absolute outcome quality for this single case, not relative comparison to other patients.

---

Additional guidance:

- Use SOFA score and ICU length of stay only as contextual information, not as primary determinants.
- Evaluate trajectory, reversibility of organ failure, and final disposition over isolated events.
- Explicit hospice/CMO/DNR with non-escalation intent strongly favors Score 3.

### Dialysis Trajectory Guidance (IMPORTANT)

Dialysis status at ICU admission and discharge may be provided.

Interpret dialysis in a trajectory-based manner:

- If dialysis was already present at ICU admission and continues at discharge, this likely reflects chronic baseline ESRD and should NOT automatically lower the score.
- If dialysis is NEW during this hospitalization and persists at discharge, this represents new major organ failure and strongly favors Score 3.
- If dialysis was required transiently during ICU stay but discontinued before discharge, consider this partial recovery rather than permanent failure.
- Dialysis alone should not determine score; consider overall reversibility, baseline status, and global trajectory.
- Chronic ESRD at baseline does not by itself imply poor ICU treatment quality.

### Discharge Location Guidance

Discharge location may be provided (e.g., HOME, HOME HEALTH CARE, REHAB, SKILLED NURSING FACILITY, LONG TERM ACUTE CARE, HOSPICE).

Interpret disposition in a trajectory-based manner:

- Discharge to HOME or HOME HEALTH CARE generally reflects good functional recovery.
- Discharge to REHAB or SKILLED NURSING FACILITY often reflects recovery with residual functional limitations but does NOT imply catastrophic outcome.
- LONG TERM ACUTE CARE may indicate persistent organ support needs or prolonged critical illness.
- HOSPICE or death reflects end-of-life trajectory.

Discharge location alone should NOT determine the score. Always interpret disposition together with organ failure trajectory, reversibility, and overall clinical course.

---

### IMPORTANT: WHEN NOT TO ASSIGN A SCORE

If the discharge summary is missing, nonsensical, corrupted, empty, or lacks sufficient clinical information to reasonably assess ICU course and outcome trajectory, do NOT guess.

Use Score = 0 only if clinical course and outcome cannot be reasonably inferred.

In such cases:
- Assign Score = 0
- Assign Task Relevance Score = 0.00
- Provide a brief justification stating why outcome assessment is not possible.

---

Task Relevance Score Guidance:

Task relevance reflects how informative the case is for evaluating sepsis-related management, not how clear or extreme the outcome appears.

Task relevance scores must be between 0.00 and 1.00.

Use high task relevance scores (>=0.90) only when:
- The discharge summary clearly describes sepsis or shock management.
- Hemodynamic course (e.g., fluids, vasopressors, blood pressure, organ perfusion) is well documented.
- The outcome is explicitly supported by documented hemodynamic course.

Reduce task relevance scores when:
- Sepsis management is minimally described or unclear.
- The ICU course is dominated by unrelated conditions.
- Hemodynamic interventions are not clearly documented.
- The outcome cannot be clearly attributed to sepsis-related processes.
- The outcome is clear but the role of sepsis management is not clearly described.

Use lower task relevance scores (<0.75) when relevance is limited, and <0.50 when the case provides little useful information about sepsis treatment.

Do NOT assign high task relevance scores based solely on discharge disposition or survival status without supporting ICU course details.

Do NOT assign high task relevance scores if sepsis management is not clearly described.

Task:

You will receive a structured header followed by the discharge summary narrative.

If sufficient information is present:
- Assign an outcome score from 1-5 and a task relevance score from 0.00-1.00.
- Provide a concise justification summarizing the single dominant clinical factor that most strongly determined the outcome category.

The justification must:
- Mention at most ONE dominant factor.
- Be 1-2 sentences only, no more than 40 words.
- Use plain clinical language.
- Avoid uncertainty hedging.
- Avoid restating the score definition verbatim.
- Avoid patient identifiers.

STRICT OUTPUT FORMAT:

Output EXACTLY TWO LINES.

Line 1:

<SCORE>,<TASK_RELEVANCE_SCORE>

Line 2:

<JUSTIFICATION>

No extra text before or after.

Now here is the discharge summary to evaluate:

{note}
]
\end{lstlisting}

\subsubsection{Example: Narrative-Derived Trajectory Quality Signal}

\noindent
\textbf{Representative Discharge Summary (Excerpt)}  
\begin{quote}
\ttfamily
Patient presented with hypotension and hypoglycemia requiring brief vasopressor support.  
He was admitted to the coronary care unit and rapidly stabilized, with norepinephrine successfully weaned shortly after admission. \\

Cardiac catheterization revealed multivessel coronary artery disease, and two stents were placed with good procedural outcome.  
His medical therapy was subsequently optimized, including adjustment of heart failure and diabetes management. \\

Following stabilization, the patient remained hemodynamically stable without further need for pressors.  
He was discharged home with outpatient follow-up planned for additional coronary intervention.
\end{quote}

\vspace{0.5em}
\noindent
\textbf{Trajectory Quality Score (TQS):} 4 \\
\textbf{Task Relevance Score:} 0.82 \\

\vspace{0.5em}
\noindent
\textbf{LLM-Generated Justification:}  
\begin{quote}
\itshape
Rapid weaning off norepinephrine after brief pressor support and successful coronary revascularization enabled discharge home with only chronic heart failure limitations.
\end{quote}

\subsection{NEWS2-Based Reward Function}

We implemented a reward function based on the National Early Warning Score 2 (NEWS2). The NEWS2 score is a composite clinical severity metric derived from routinely measured physiological variables.

At each time step $t$, the NEWS2 score is computed from the \textit{next-state} physiological variables:
\[
\text{NEWS2}_{t+1} = 
s_{\text{RR}} + s_{\text{SpO}_2} + s_{\text{O}_2} + s_{\text{BP}} + s_{\text{HR}} + s_{\text{Temp}} + s_{\text{CNS}},
\]
where each component score corresponds to discretized clinical measurements:
\begin{itemize}
    \item Respiratory rate (RR)
    \item Oxygen saturation (SpO$_2$)
    \item Supplemental oxygen requirement (mechanical ventilation)
    \item Systolic blood pressure (BP)
    \item Heart rate (HR)
    \item Temperature (Temp)
    \item Level of consciousness (CNS), approximated using $GCS < 15$
\end{itemize}

Each component is assigned a score according to standard NEWS2 thresholds, resulting in a total score ranging from 0 (normal) to a theoretical maximum of 20 (severe physiological derangement).

To ensure comparability across time steps, the NEWS2 score is normalized:
\[
\widetilde{\text{NEWS2}}_{t+1} = \frac{\text{NEWS2}_{t+1}}{20}.
\]

The reward at each time step is then defined as the negative normalized NEWS2 score:
\[
r_t = - \widetilde{\text{NEWS2}}_{t+1}.
\]

This formulation encourages transitions toward physiologically stable states by assigning higher (less negative) rewards to lower NEWS2 scores.

At terminal time steps, the reward is overridden to reflect final outcomes:
\[
r_T =
\begin{cases}
-1, & \text{if in-hospital mortality occurs}, \\
0, & \text{if the patient survives}.
\end{cases}
\]

This terminal adjustment ensures that long-term outcomes are incorporated, while intermediate rewards capture short-term physiological stability.

Overall, the NEWS2-based reward combines dense intermediate feedback with a sparse terminal signal, providing a structured yet heuristic objective for policy learning.

\subsection{Outcome Metric Definitions}
\label{app:outcomes}

All outcome metrics were computed using a unified discretization of patient trajectories into fixed 4-hour intervals, consistent with the reinforcement learning time step. For each ICU stay, a complete time grid was constructed from ICU admission ($t=0$) up to 28 days.

At each time step $t$, patient status was defined based on (i) survival status, (ii) organ support usage, and (iii) hemodynamic measurements.

\subsubsection{Organ Support--Free Days (OSFD)}

Organ support--free days quantify the number of days during which a patient is both alive and free from major organ support therapies.

We define organ support at time $t$ as the presence of any of the following:
\begin{itemize}
    \item vasopressor administration,
    \item mechanical ventilation, or
    \item renal replacement therapy (including CRRT).
\end{itemize}

At each time step $t$, we define:
\begin{equation}
\text{SupportFreeAlive}_t = 
\mathbf{1} \big( \text{alive}_t = 1 \;\land\; \text{no support}_t \big),
\end{equation}
where $\mathbf{1}(\cdot)$ denotes the indicator function.

The OSFD metric over a given time horizon is computed as:
\begin{equation}
\text{OSFD} = \frac{1}{6} \sum_{t=1}^{T} \text{SupportFreeAlive}_t,
\end{equation}
where each day corresponds to six 4-hour intervals.

We report OSFD-7, defined over the first 7 days (168 hours) following ICU admission.

\subsubsection{Time to Shock Resolution (TSR)}

Shock at time $t$ is defined as:
\begin{equation}
\text{Shock}_t =
\begin{cases}
1, & \text{if vasopressor support is present} \\
1, & \text{if } \text{MAP}_t < 65 \text{ mmHg} \\
0, & \text{otherwise}
\end{cases}
\end{equation}
where $\text{MAP}_t$ denotes the mean arterial pressure at time $t$.

Shock-free status is defined as:
\begin{equation}
\text{ShockFree}_t = \mathbf{1}(\text{Shock}_t = 0).
\end{equation}

Time to shock resolution (TSR) measures the time required to achieve sustained hemodynamic stability. Sustained resolution is defined as remaining shock-free for 12 consecutive hours:
\begin{equation}
n = \frac{12\ \text{hours}}{4\ \text{hours per interval}} = 3\ \text{intervals}.
\end{equation}

TSR is defined as the earliest time step $t$ such that:
\begin{equation}
\text{TSR} = \min \left\{ t \;:\; \sum_{k=t}^{t+n-1} \text{ShockFree}_k = n \right\}.
\end{equation}

TSR is reported in hours:
\begin{equation}
\text{TSR (hours)} = \frac{\text{offset}_t}{60},
\end{equation}
where $\text{offset}_t$ denotes minutes since ICU admission.

Special cases:
\begin{itemize}
    \item TSR is undefined for patients who never experience shock.
    \item If shock is not resolved within 72 hours, TSR is capped at 72 hours.
\end{itemize}

\subsubsection{Treatment Burden}

Treatment burden quantifies the intensity of administered therapies over a patient trajectory. 
To account for variation in trajectory length, we define burden as the mean treatment level per time step.

The IV fluid burden is defined as:
\begin{equation}
\text{IV Fluids Burden} = \frac{1}{T} \sum_{t=1}^{T} a^{\text{IV}}_t,
\end{equation}

and the vasopressor burden is defined as:
\begin{equation}
\text{Vasopressor Burden} = \frac{1}{T} \sum_{t=1}^{T} a^{\text{vaso}}_t,
\end{equation}

where $a^{\text{IV}}_t$ and $a^{\text{vaso}}_t$ denote the discretized treatment levels of intravenous fluids and vasopressors at time step $t$, and $T$ is the number of time steps in the trajectory.

These normalized metrics capture treatment intensity independent of trajectory duration, enabling fair comparison across patients with differing lengths of stay. Separating IV fluids and vasopressors further distinguishes volume resuscitation from vasopressor-driven hemodynamic support.

\subsubsection{Discharge Disposition Score}

Discharge disposition was used as a proxy for functional recovery and longer-term prognosis beyond survival. Each patient was assigned a discrete score based on their hospital discharge destination, reflecting increasing levels of independence and recovery.

Formally, the discharge disposition score is defined as:
\begin{equation}
\text{Discharge Score} \in \{0, \dots, 7\},
\end{equation}
with higher values indicating more favorable outcomes.

Scores were assigned according to the following mapping:
\begin{itemize}
    \item \textbf{7:} Discharge to home
    \item \textbf{6:} Discharge to home with healthcare services
    \item \textbf{5:} Discharge to rehabilitation facility
    \item \textbf{4:} Discharge to assisted living
    \item \textbf{3:} Discharge to skilled nursing facility
    \item \textbf{2:} Discharge to long-term acute care or healthcare facility
    \item \textbf{1:} Transfer to acute hospital
    \item \textbf{0:} Hospice or in-hospital death
\end{itemize}

Discharge categories that do not clearly map to a recovery trajectory, including \textit{other facility}, \textit{against medical advice}, and \textit{psychiatric facility}, were excluded from this metric and treated as missing values.

This ordering reflects a clinically motivated hierarchy of recovery, where discharge to home represents near-complete functional recovery, while lower scores correspond to increasing levels of dependency and ongoing care needs. Hospice and death are grouped as the lowest category, as both represent trajectories with minimal likelihood of meaningful recovery.

By incorporating discharge disposition into evaluation, this metric captures distinctions in recovery quality that are not reflected by mortality alone, enabling a more nuanced assessment of treatment effectiveness and longer-term patient outcomes. \\


These metrics capture:
\begin{itemize}
    \item Survival and recovery (OSFD)
    \item Hemodynamic stabilization (TSR)
    \item Treatment intensity (burden)
    \item Functional recovery (disposition)
\end{itemize}

\section{Supplementary Tables}

\begin{table}[ht]
\centering
\caption{Cohort characteristics for the MIMIC-IV and AmsterdamUMCdb sepsis cohorts.}
\label{tab:cohort_characteristics}
\resizebox{\textwidth}{!}{
\begin{threeparttable}

\begin{tabular}{llcccc}
\toprule
\textbf{Dataset} & \textbf{Cohort} & \textbf{\% Female} & \textbf{Age (years)} & \textbf{ICU Stay (hours)} & \textbf{Total (n)} \\
 &  &  & \textbf{Median (IQR)} & \textbf{Median (IQR)} &  \\
\midrule

\multirow{3}{*}{MIMIC-IV}
& Overall 
& 42.5 
& 68 (57--79) 
& 63.4 (34.7--123.2) 
& 25{,}370 \\

& Non-Survivors 
& 44.0 
& 72 (61--82) 
& 97.1 (46.2--189.4) 
& 3{,}877 \\

& Survivors 
& 42.2 
& 68 (56--78) 
& 58.4 (33.2--113.3) 
& 21{,}493 \\

\midrule

\multirow{3}{*}{AmsterdamUMCdb}
& Overall 
& 40.1 
& 65 (55--75)\textsuperscript{*} 
& 99.8 (39.3--270.8) 
& 4{,}404 \\

& Non-Survivors 
& 41.0 
& 65 (55--75)\textsuperscript{*} 
& 94.9 (39.3--278.5) 
& 966 \\

& Survivors 
& 40.0 
& 65 (55--75)\textsuperscript{*} 
& 103.9 (39.2--267.0) 
& 3{,}438 \\

\bottomrule
\end{tabular}

\begin{tablenotes}
\footnotesize
\item[*] Age was available only as categorical ranges in AmsterdamUMCdb; accordingly, patient age was approximated using the midpoint of each age category (e.g., 18--39 $\rightarrow$ 30, 40--49 $\rightarrow$ 45).
\end{tablenotes}

\end{threeparttable}
}
\end{table}

\begin{table}[ht]
\centering
\caption{Action Quantile Thresholds (4-hourly doses)}
\resizebox{\textwidth}{!}{
\begin{tabular}{lccccc}
\toprule
\textbf{Quantile} & \textbf{0} & \textbf{1} & \textbf{2} & \textbf{3} & \textbf{4} \\
\midrule
\textbf{IV Fluids (ml)} & 0 & 0.00--50.05 & 50.05--213.33 & 213.33--520.0 & $>$520.0 \\
\textbf{Vasopressors (mcg/kg)} & 0 & 0.00--7.20 & 7.20--17.41 & 17.41--40.06 & $>$40.06 \\
\bottomrule
\end{tabular}
}
\label{tab:quantile_thresholds}
\end{table}

\begin{table}[t]
\centering
\caption{
Agreement between clinician-assigned and LLM-derived trajectory quality scores (TQS) and task relevance (TR) scores on a subset of 150 discharge summaries.
}
\label{tab:annotation_agreement}
\resizebox{\textwidth}{!}{
\small
\begin{tabular}{llcccc}
\toprule
\textbf{Task} & \textbf{Comparison} & \textbf{Quadratic-weighted $\kappa$} & \textbf{Spearman $\rho$} & \textbf{Exact Agreement} & \textbf{Within-1 Agreement} \\
\midrule

\multirow{3}{*}{TQS}
& Clinician--clinician pairwise 
& $0.917 \pm 0.007$ 
& $0.931 \pm 0.006$ 
& $0.736$ 
& $0.998$ \\

& LLM--clinician pairwise 
& $0.910 \pm 0.023$ 
& $0.923 \pm 0.011$ 
& $0.722$ 
& $0.989$ \\

& LLM--clinician consensus 
& $0.915$ 
& $0.926$ 
& $0.747$ 
& $0.987$ \\

\midrule

\multirow{3}{*}{TR}
& Clinician--clinician pairwise 
& $0.684 \pm 0.041$ 
& $0.742 \pm 0.028$ 
& $0.518$ 
& $0.821$ \\

& LLM--clinician pairwise 
& $0.661 \pm 0.052$ 
& $0.718 \pm 0.036$ 
& $0.487$ 
& $0.803$ \\

& LLM--clinician consensus 
& $0.701$ 
& $0.761$ 
& $0.536$ 
& $0.834$ \\

\bottomrule
\end{tabular}
}
\vspace{0.5em}

\begin{minipage}{0.96\linewidth}
\footnotesize
\textbf{Notes:}
TQS = trajectory quality score; TR = task relevance. Clinician consensus scores were computed using the median score across three clinicians. Values are reported as mean $\pm$ standard deviation across pairwise comparisons where applicable. Within-1 agreement denotes predictions differing by at most one ordinal category.
\end{minipage}

\end{table}

\vspace{0.5em}

\begin{table}[t]
\centering
\caption{
External validation on AmsterdamUMCdb: Spearman correlation between trajectory-level cumulative rewards and clinical outcomes for different reward formulations.
}
\label{tab:reward_correlations_amsterdam}
\resizebox{\columnwidth}{!}{
\begin{tabular}{lccccc}
\toprule
\textbf{Reward} & \textbf{OSFD-7} & \textbf{TSR} & \textbf{Mortality} & \textbf{IV Fluid Burden} & \textbf{Vasopressor Burden} \\
\midrule

CN-PR  
& \textbf{0.74} 
& \textbf{-0.36} 
& -0.69 
& \textbf{-0.17} 
& \textbf{-0.45} \\

CN-PR (No TR) 
& 0.72 
& -0.31 
& -0.68 
& -0.15 
& -0.43 \\

Mortality-only 
& 0.68 
& -0.27 
& \textbf{-0.76} 
& -0.04 
& -0.11 \\

SOFA-Lac 
& 0.67 
& -0.20 
& -0.72 
& -0.04 
& -0.11 \\

NEWS2 
& 0.60 
& -0.26 
& -0.56 
& -0.11 
& -0.34 \\

Permuted-TQS
& -0.04 
& 0.04 
& 0.06 
& -0.06 
& 0.06 \\

\bottomrule
\end{tabular}
}
\vspace{0.3em}

\footnotesize
\textit{Notes:} OSFD-7 = organ support--free days at 7 days; TSR = time to shock resolution; TR = task-relevance weighting. Bold values indicate the strongest association for each outcome. Discharge disposition score was unavailable in the AmsterdamUMCdb dataset.
\end{table}

\begin{table}[ht]
\centering
\scriptsize
\setlength{\tabcolsep}{-0.5pt}
\renewcommand{\arraystretch}{1.05}

\caption{Multivariable regression: MIMIC-IV. Coefficients shown as $\beta$ (95\% CI).}
\label{tab:regression}
\resizebox{\textwidth}{!}{
\begin{tabular}{lccccc}
\toprule
 & CN-PR & SOFA-Lac policy & Mortality policy & NEWS2 policy & Random policy \\
\midrule
\multicolumn{6}{l}{\textbf{A. Hospital Mortality}} \\
Distance & 0.48 (0.4, 0.55)*** & 0.5 (0.42, 0.58)*** & 0.5 (0.42, 0.58)*** & 0.31 (0.24, 0.38)*** & -0.01 (-0.09, 0.06) \\
Age & 0.02 (0.02, 0.03)*** & 0.02 (0.02, 0.03)*** & 0.02 (0.02, 0.03)*** & 0.02 (0.02, 0.03)*** & 0.02 (0.02, 0.03)*** \\
SOFA & 0.11 (0.08, 0.14)*** & 0.11 (0.08, 0.13)*** & 0.11 (0.08, 0.14)*** & 0.12 (0.1, 0.15)*** & 0.15 (0.12, 0.17)*** \\
MechVent & 0.18 (-0.03, 0.39) & 0.1 (-0.11, 0.31) & 0.09 (-0.12, 0.3) & 0.14 (-0.07, 0.34) & 0.11 (-0.1, 0.31) \\
S.I. & 0.75 (0.44, 1.06)*** & 0.73 (0.41, 1.04)*** & 0.66 (0.34, 0.97)*** & 0.88 (0.58, 1.19)*** & 0.99 (0.69, 1.29)*** \\
Lactate & 0.23 (0.19, 0.28)*** & 0.23 (0.18, 0.27)*** & 0.23 (0.19, 0.27)*** & 0.25 (0.2, 0.29)*** & 0.25 (0.21, 0.3)*** \\
Elixhauser & 0.02 (0.02, 0.03)*** & 0.02 (0.02, 0.03)*** & 0.03 (0.02, 0.03)*** & 0.02 (0.02, 0.03)*** & 0.03 (0.02, 0.03)*** \\
\midrule
\multicolumn{6}{l}{\textbf{B. OSFD-7}} \\
Distance & -0.84 (-0.9, -0.79)*** & -0.64 (-0.7, -0.58)*** & -0.62 (-0.68, -0.56)*** & -0.49 (-0.55, -0.43)*** & 0.21 (0.15, 0.26)*** \\
Age & 0.0 (0.0, 0.01)** & 0.01 (0.0, 0.01)** & 0.0 (0.0, 0.01)* & 0.0 (0.0, 0.01)* & 0.0 (-0.0, 0.01) \\
SOFA & -0.18 (-0.2, -0.16)*** & -0.2 (-0.22, -0.17)*** & -0.2 (-0.22, -0.18)*** & -0.21 (-0.23, -0.19)*** & -0.24 (-0.27, -0.22)*** \\
MechVent & 0.06 (-0.08, 0.21) & 0.15 (0.0, 0.29)* & 0.05 (-0.1, 0.2) & 0.11 (-0.04, 0.26) & 0.15 (-0.0, 0.3) \\
S.I. & -0.26 (-0.51, -0.01)* & -0.37 (-0.61, -0.12)** & -0.4 (-0.66, -0.15)** & -0.52 (-0.77, -0.26)*** & -0.74 (-0.99, -0.49)*** \\
Lactate & -0.2 (-0.24, -0.17)*** & -0.2 (-0.24, -0.17)*** & -0.22 (-0.26, -0.19)*** & -0.23 (-0.27, -0.2)*** & -0.25 (-0.29, -0.21)*** \\
Elixhauser & -0.03 (-0.03, -0.02)*** & -0.02 (-0.03, -0.02)*** & -0.03 (-0.03, -0.02)*** & -0.03 (-0.03, -0.02)*** & -0.03 (-0.03, -0.02)*** \\
\midrule
\multicolumn{6}{l}{\textbf{C. Time to Shock Resolution (hours)}} \\
Distance & 7.79 (7.05, 8.54)*** & 5.35 (4.58, 6.12)*** & 5.12 (4.36, 5.89)*** & 6.46 (5.71, 7.22)*** & -4.17 (-4.84, -3.5)*** \\
Age & 0.09 (0.05, 0.13)*** & 0.08 (0.04, 0.12)*** & 0.08 (0.04, 0.12)*** & 0.09 (0.05, 0.13)*** & 0.08 (0.04, 0.12)*** \\
SOFA & 1.71 (1.45, 1.98)*** & 1.96 (1.68, 2.23)*** & 1.99 (1.72, 2.27)*** & 1.91 (1.64, 2.19)*** & 2.22 (1.95, 2.49)*** \\
MechVent & -2.76 (-4.38, -1.14)*** & -3.9 (-5.57, -2.22)*** & -3.91 (-5.59, -2.23)*** & -3.25 (-4.9, -1.59)*** & -4.09 (-5.77, -2.42)*** \\
S.I. & 7.64 (4.62, 10.67)*** & 8.98 (5.86, 12.11)*** & 8.66 (5.6, 11.73)*** & 9.72 (6.8, 12.64)*** & 11.73 (8.7, 14.77)*** \\
Lactate & 0.63 (0.23, 1.02)** & 0.64 (0.22, 1.07)** & 0.64 (0.21, 1.06)** & 0.88 (0.48, 1.28)*** & 1.09 (0.67, 1.5)*** \\
Elixhauser & 0.06 (-0.02, 0.13) & 0.07 (-0.01, 0.14) & 0.08 (0.0, 0.15)* & 0.06 (-0.02, 0.13) & 0.1 (0.02, 0.17)* \\
\midrule
\multicolumn{6}{l}{\textbf{D. Discharge Disposition Score}} \\
Distance & -0.47 (-0.53, -0.41)*** & -0.38 (-0.44, -0.32)*** & -0.33 (-0.39, -0.27)*** & -0.3 (-0.36, -0.24)*** & 0.06 (0.01, 0.12)* \\
Age & -0.04 (-0.05, -0.04)*** & -0.04 (-0.05, -0.04)*** & -0.04 (-0.05, -0.04)*** & -0.04 (-0.05, -0.04)*** & -0.04 (-0.05, -0.04)*** \\
SOFA & -0.1 (-0.13, -0.08)*** & -0.11 (-0.13, -0.09)*** & -0.12 (-0.14, -0.1)*** & -0.12 (-0.14, -0.1)*** & -0.14 (-0.16, -0.12)*** \\
MechVent & 0.15 (-0.01, 0.32) & 0.2 (0.04, 0.36)* & 0.15 (-0.01, 0.32) & 0.18 (0.01, 0.34)* & 0.2 (0.04, 0.37)* \\
S.I. & -0.62 (-0.86, -0.39)*** & -0.67 (-0.91, -0.43)*** & -0.73 (-0.97, -0.49)*** & -0.76 (-1.0, -0.52)*** & -0.9 (-1.14, -0.66)*** \\
Lactate & -0.17 (-0.2, -0.14)*** & -0.17 (-0.2, -0.14)*** & -0.19 (-0.22, -0.16)*** & -0.19 (-0.22, -0.16)*** & -0.2 (-0.23, -0.16)*** \\
Elixhauser & -0.03 (-0.04, -0.03)*** & -0.03 (-0.04, -0.02)*** & -0.03 (-0.04, -0.03)*** & -0.03 (-0.04, -0.03)*** & -0.03 (-0.04, -0.03)*** \\
\midrule
\multicolumn{6}{l}{\textbf{E. IV Fluid Burden (mL)}} \\
Distance & 60.74 (52.69, 68.8)*** & 16.83 (8.37, 25.28)*** & 17.75 (9.42, 26.07)*** & 32.0 (23.85, 40.15)*** & 14.5 (7.06, 21.95)*** \\
Age & -2.43 (-2.87, -1.99)*** & -2.37 (-2.82, -1.92)*** & -2.37 (-2.82, -1.92)*** & -2.37 (-2.81, -1.92)*** & -2.24 (-2.69, -1.8)*** \\
SOFA & 4.77 (1.72, 7.83)** & 8.23 (5.16, 11.3)*** & 8.21 (5.14, 11.28)*** & 7.03 (3.95, 10.11)*** & 10.29 (7.34, 13.24)*** \\
MechVent & 19.43 (-0.24, 39.1) & 14.1 (-5.49, 33.7) & 13.89 (-5.67, 33.45) & 16.58 (-3.05, 36.21) & 13.88 (-5.59, 33.35) \\
S.I. & 135.93 (100.25, 171.61)*** & 163.29 (127.01, 199.58)*** & 161.36 (125.01, 197.71)*** & 158.2 (122.52, 193.88)*** & 174.84 (139.27, 210.41)*** \\
Lactate & 33.72 (26.83, 40.62)*** & 35.79 (28.93, 42.64)*** & 35.82 (28.95, 42.68)*** & 36.37 (29.63, 43.11)*** & 35.88 (28.98, 42.78)*** \\
Elixhauser & -0.13 (-0.95, 0.69) & -0.03 (-0.87, 0.8) & -0.0 (-0.84, 0.83) & -0.1 (-0.93, 0.73) & 0.02 (-0.81, 0.86) \\
\midrule
\multicolumn{6}{l}{\textbf{F. Vasopressor Burden (mcg)}} \\
Distance & 4.86 (4.21, 5.5)*** & 3.8 (3.09, 4.51)*** & 3.57 (2.9, 4.25)*** & 1.83 (1.14, 2.53)*** & 0.9 (0.33, 1.46)** \\
Age & -0.02 (-0.05, 0.01) & -0.02 (-0.05, 0.01) & -0.02 (-0.05, 0.01) & -0.01 (-0.04, 0.02) & -0.0 (-0.03, 0.03) \\
SOFA & 1.78 (1.5, 2.05)*** & 1.87 (1.59, 2.15)*** & 1.9 (1.62, 2.19)*** & 2.05 (1.75, 2.34)*** & 2.24 (1.95, 2.52)*** \\
MechVent & -0.1 (-1.68, 1.47) & -0.59 (-2.18, 1.01) & -0.63 (-2.22, 0.96) & -0.45 (-2.07, 1.16) & -0.59 (-2.21, 1.03) \\
S.I. & 9.39 (6.55, 12.24)*** & 9.93 (7.06, 12.8)*** & 9.79 (6.92, 12.66)*** & 11.35 (8.46, 14.23)*** & 12.26 (9.39, 15.12)*** \\
Lactate & 3.36 (2.49, 4.23)*** & 3.36 (2.47, 4.26)*** & 3.39 (2.5, 4.28)*** & 3.54 (2.66, 4.42)*** & 3.51 (2.63, 4.38)*** \\
Elixhauser & 0.04 (-0.03, 0.1) & 0.03 (-0.03, 0.1) & 0.04 (-0.03, 0.1) & 0.04 (-0.02, 0.11) & 0.05 (-0.02, 0.12) \\
\\
\bottomrule
\end{tabular}
}

\begin{flushleft}
\footnotesize
\centering
\textit{Notes:} * $p < 0.05$, ** $p < 0.01$, *** $p < 0.001$. 
Intercept terms are omitted. S.I. = shock index. All models adjusted for age, SOFA score, mechanical ventilation, S.I., lactate, and Elixhauser score.
\end{flushleft}

\end{table}

\begin{table}[ht]
\centering
\scriptsize
\setlength{\tabcolsep}{-0.5pt}
\renewcommand{\arraystretch}{1.05}

\caption{Multivariable regression: AmsterdamUMCdb (External Validation). Coefficients shown as $\beta$ (95\% CI).}
\label{tab:regression2}
\resizebox{\textwidth}{!}{
\begin{tabular}{lccccc}
\toprule
 & CN-PR & SOFA-Lac policy & Mortality policy & NEWS2 policy & Random policy \\
\midrule
\multicolumn{6}{l}{\textbf{A. Hospital Mortality}} \\
Distance & 0.54 (0.42, 0.65)*** & 0.6 (0.47, 0.72)*** & 0.58 (0.47, 0.7)*** & 0.31 (0.2, 0.43)*** & 0.24 (0.12, 0.37)*** \\
Age & 0.03 (0.02, 0.04)*** & 0.03 (0.02, 0.04)*** & 0.03 (0.02, 0.04)*** & 0.03 (0.02, 0.04)*** & 0.03 (0.02, 0.04)*** \\
SOFA & 0.22 (0.18, 0.26)*** & 0.21 (0.17, 0.25)*** & 0.22 (0.18, 0.26)*** & 0.22 (0.17, 0.26)*** & 0.24 (0.19, 0.28)*** \\
MechVent & 0.33 (0.23, 0.42)*** & 0.32 (0.22, 0.42)*** & 0.31 (0.21, 0.4)*** & 0.33 (0.23, 0.42)*** & 0.32 (0.22, 0.41)*** \\
S.I. & 0.42 (0.06, 0.78)* & 0.41 (0.05, 0.78)* & 0.42 (0.05, 0.79)* & 0.49 (0.13, 0.86)** & 0.48 (0.12, 0.83)** \\
Lactate & 0.17 (0.13, 0.22)*** & 0.18 (0.14, 0.22)*** & 0.18 (0.13, 0.22)*** & 0.18 (0.14, 0.23)*** & 0.18 (0.13, 0.22)*** \\
\midrule
\multicolumn{6}{l}{\textbf{B. OSFD-7}} \\
Distance & -0.59 (-0.68, -0.5)*** & -0.47 (-0.56, -0.38)*** & -0.52 (-0.61, -0.43)*** & -0.34 (-0.43, -0.25)*** & 0.04 (-0.06, 0.14) \\
Age & -0.01 (-0.02, -0.01)*** & -0.01 (-0.02, -0.01)*** & -0.01 (-0.02, -0.01)*** & -0.01 (-0.02, -0.0)*** & -0.01 (-0.01, -0.0)** \\
SOFA & -0.34 (-0.37, -0.31)*** & -0.35 (-0.38, -0.32)*** & -0.34 (-0.37, -0.31)*** & -0.36 (-0.39, -0.33)*** & -0.38 (-0.41, -0.35)*** \\
MechVent & -0.72 (-0.78, -0.66)*** & -0.72 (-0.78, -0.66)*** & -0.72 (-0.78, -0.66)*** & -0.73 (-0.79, -0.67)*** & -0.71 (-0.77, -0.65)*** \\
S.I. & -0.3 (-0.54, -0.07)* & -0.27 (-0.51, -0.03)* & -0.17 (-0.42, 0.07) & -0.27 (-0.51, -0.03)* & -0.31 (-0.55, -0.06)* \\
Lactate & -0.04 (-0.08, -0.01)** & -0.05 (-0.08, -0.01)** & -0.06 (-0.09, -0.03)*** & -0.05 (-0.09, -0.02)** & -0.06 (-0.1, -0.03)*** \\
\midrule
\multicolumn{6}{l}{\textbf{C. Time to Shock Resolution (hours)}} \\
Distance & 5.64 (4.34, 6.94)*** & 4.22 (2.93, 5.51)*** & 4.44 (3.13, 5.74)*** & 2.45 (1.12, 3.77)*** & -3.27 (-4.53, -2.02)*** \\
Age & 0.07 (-0.0, 0.15) & 0.07 (-0.01, 0.14) & 0.07 (-0.0, 0.15) & 0.06 (-0.01, 0.14) & 0.05 (-0.03, 0.12) \\
SOFA & 2.1 (1.66, 2.53)*** & 2.21 (1.76, 2.65)*** & 2.22 (1.78, 2.66)*** & 2.33 (1.89, 2.77)*** & 2.4 (1.97, 2.83)*** \\
MechVent & 5.76 (4.93, 6.58)*** & 5.71 (4.88, 6.54)*** & 5.75 (4.91, 6.59)*** & 5.73 (4.88, 6.58)*** & 5.39 (4.57, 6.21)*** \\
S.I. & 4.05 (0.26, 7.84)* & 6.0 (2.27, 9.74)** & 5.85 (2.25, 9.45)** & 5.68 (1.94, 9.43)** & 6.22 (2.44, 9.99)** \\
Lactate & 0.03 (-0.42, 0.48) & -0.04 (-0.48, 0.41) & -0.05 (-0.5, 0.39) & 0.02 (-0.43, 0.48) & 0.25 (-0.19, 0.7) \\
\midrule
\multicolumn{6}{l}{\textbf{D. IV Fluid Burden (mL)}} \\
Distance & 50.17 (36.26, 64.08)*** & 21.6 (7.39, 35.82)** & 18.49 (5.76, 31.21)** & 44.25 (29.47, 59.03)*** & 2.42 (-12.9, 17.75) \\
Age & -0.49 (-1.23, 0.25) & -0.55 (-1.29, 0.2) & -0.57 (-1.32, 0.19) & -0.51 (-1.26, 0.23) & -0.61 (-1.36, 0.15) \\
SOFA & -0.6 (-5.03, 3.83) & 0.66 (-3.92, 5.23) & 0.8 (-3.74, 5.34) & -0.07 (-4.52, 4.37) & 1.91 (-2.57, 6.39) \\
MechVent & 16.73 (8.86, 24.6)*** & 17.44 (9.34, 25.54)*** & 17.62 (9.51, 25.73)*** & 19.34 (11.62, 27.06)*** & 18.35 (10.32, 26.38)*** \\
S.I. & 19.08 (-28.21, 66.37) & 23.84 (-23.05, 70.74) & 25.34 (-21.35, 72.03) & 22.2 (-24.23, 68.62) & 29.19 (-18.39, 76.76) \\
Lactate & 31.76 (22.02, 41.5)*** & 32.69 (22.82, 42.56)*** & 32.76 (22.92, 42.61)*** & 31.56 (21.65, 41.46)*** & 33.05 (23.18, 42.91)*** \\
\midrule
\multicolumn{6}{l}{\textbf{E. Vasopressor Burden (mcg)}} \\
Distance & 7.14 (4.35, 9.93)*** & 4.18 (1.43, 6.94)** & 5.4 (2.63, 8.16)*** & 4.61 (1.64, 7.57)** & 0.8 (-2.08, 3.68) \\
Age & 0.36 (0.23, 0.48)*** & 0.34 (0.21, 0.46)*** & 0.34 (0.21, 0.46)*** & 0.34 (0.21, 0.47)*** & 0.33 (0.2, 0.46)*** \\
SOFA & 4.56 (3.62, 5.49)*** & 4.78 (3.85, 5.71)*** & 4.65 (3.69, 5.62)*** & 4.84 (3.94, 5.75)*** & 5.03 (4.11, 5.96)*** \\
MechVent & 5.45 (4.38, 6.51)*** & 5.44 (4.36, 6.51)*** & 5.36 (4.29, 6.44)*** & 5.51 (4.43, 6.6)*** & 5.43 (4.33, 6.53)*** \\
S.I. & 26.89 (16.04, 37.74)*** & 27.13 (16.3, 37.96)*** & 26.94 (16.08, 37.81)*** & 26.88 (16.06, 37.69)*** & 27.07 (16.22, 37.93)*** \\
Lactate & 6.4 (4.61, 8.2)*** & 6.5 (4.68, 8.31)*** & 6.44 (4.65, 8.23)*** & 6.49 (4.72, 8.25)*** & 6.6 (4.82, 8.38)*** \\
\\
\bottomrule
\end{tabular}
}
\begin{flushleft}
\footnotesize
\centering
\textit{Notes:} * $p < 0.05$, ** $p < 0.01$, *** $p < 0.001$. 
Intercept terms are omitted. S.I. = shock index. All models adjusted for age, SOFA score, mechanical ventilation, S.I., and lactate.
\end{flushleft}
\end{table}

\FloatBarrier
\section{Supplementary Figures}

\begin{figure}[htbp]
\centering
\includegraphics[width=0.8\textwidth]{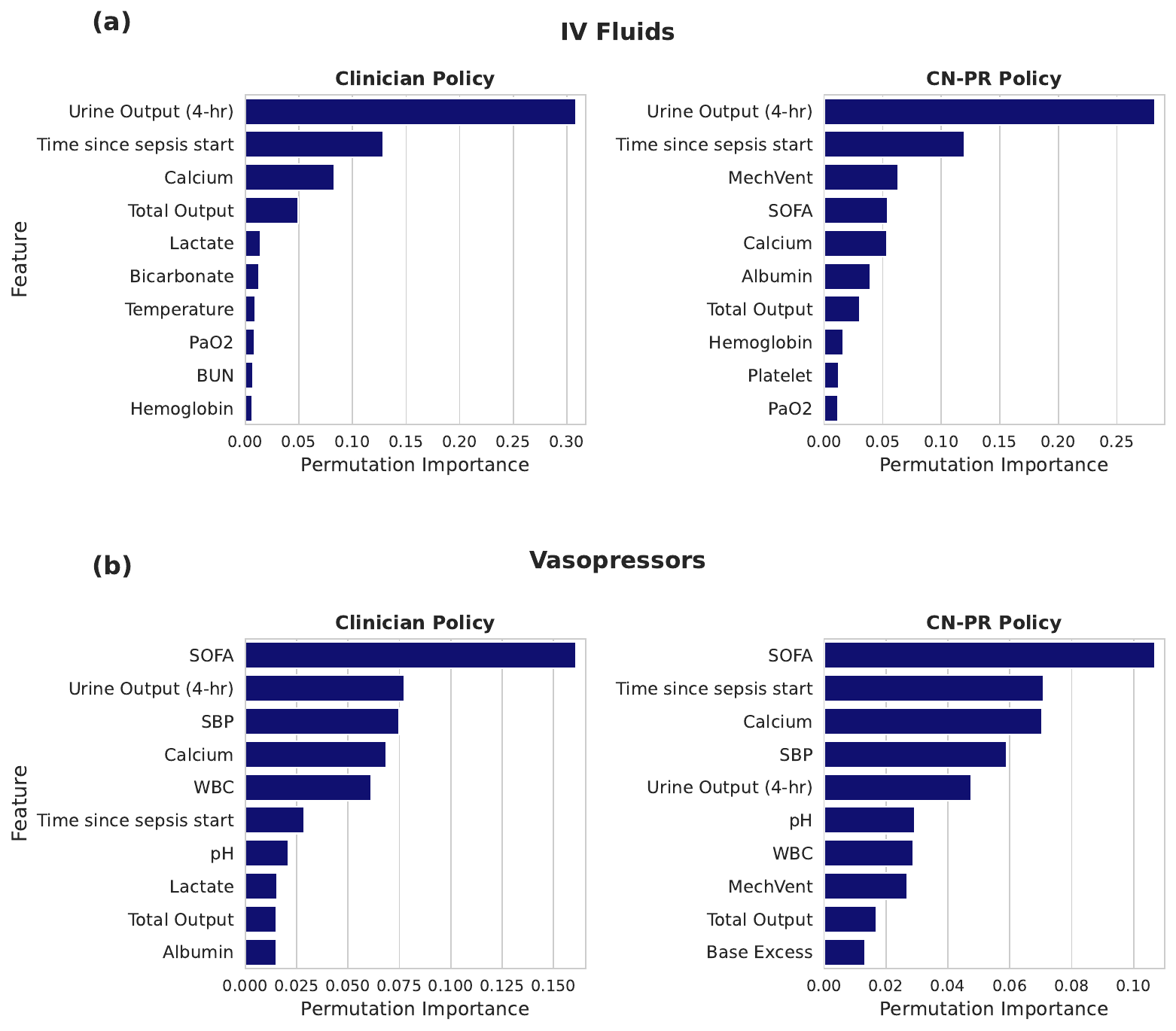} 
\caption{
Permutation feature importance for predicting discretized treatment actions under clinician and CN-PR policies. 
Separate models were trained for (a) IV fluids and (b) vasopressors, with importance computed via permutation. 
The top 10 features for each policy are shown to facilitate comparison of key drivers of treatment decisions.
}
\label{fig:feature_importance}
\vspace{-0.5em} 
\end{figure}

\clearpage

\end{document}